\documentclass[runningheads]{llncs}

\usepackage{eccv}

\usepackage{eccvabbrv}
\usepackage{url}
\usepackage{booktabs}
\usepackage{graphicx}
\usepackage{multirow}
\usepackage{subcaption}
\usepackage{wrapfig}
\usepackage{adjustbox}
\usepackage{enumitem}
\usepackage{algorithm}
\usepackage{algorithmic}
\usepackage{setspace}
\usepackage{bm}
\usepackage{makecell}
\usepackage{caption}
\usepackage{subcaption}

\usepackage{comment}
\usepackage{graphicx}
\usepackage{booktabs}
\usepackage{multicol}
\usepackage{threeparttable}
\usepackage{amsfonts}       
\usepackage{amssymb}     
\usepackage{amsmath}     
\usepackage{nicefrac}       
\usepackage{microtype}      
\usepackage{xcolor}         
\usepackage{bm}     
\usepackage{xfrac} 
\usepackage{array}
\usepackage[accsupp]{axessibility}  

\usepackage{hyperref}

\usepackage{orcidlink}

\begin{document}

\title{OP4KSR: One-Step Patch-Free 4K Super-Resolution with Periodic Artifact Suppression}

\titlerunning{OP4KSR}

\author{
Chengyan Deng\inst{1} \and
Pengbin Yu\inst{2} \and
Zhentao Chen\inst{2} \and
Wei Shen\inst{2} \and
Kai Zhang\inst{3} \and
Meng Li\inst{2} \and
Lunxi Yuan\inst{2} \and
Xue Zhou\inst{1} \and
Li Yu\inst{1}\thanks{Corresponding author.}
}

\authorrunning{C. Deng et al.}

\institute{
School of Automation Engineering, University of Electronic Science and Technology of China
\and
OPPO AI Center, OPPO Inc.
\and
School of Intelligence Science and Technology, Nanjing University\\
\email{dengchengyan@std.uestc.edu.cn, lyu@uestc.edu.cn}
}

\maketitle

\begin{abstract}
Diffusion-based real-world image super-resolution (Real-ISR) has achieved remarkable perceptual quality; however, directly super-resolving images to 4K remains limited by extreme memory consumption. Consequently, prior methods adopt patch-based inference, sacrificing global context and introducing semantic confusion, spatial inconsistency, and severe latency. We propose OP4KSR, a one-step patch-free 4K SR approach built upon the powerful Flux backbone. By leveraging the extreme-compression F16 VAE, OP4KSR makes 4K SR inference tractable under practical GPU budgets, preserving global spatial-semantic coherence while enabling highly efficient inference. However, adapting this one-step architecture intrinsically triggers severe periodic artifacts. We trace this to a RoPE base frequency allocation mismatch and intra-token spatial ambiguity, both exacerbated by the lack of iterative refinement. To suppress these artifacts, we couple RoPE base frequency rescaling (RFR) with an autocorrelation-based periodicity loss ($\mathcal{L}_\text{AP}$). Furthermore, we curate a dedicated training dataset alongside three benchmarks (one synthetic and two real-world) to advance 4K SR research. Extensive experiments demonstrate that OP4KSR achieves competitive perceptual quality with efficient inference, generating a $4096\times4096$ output in only 5.75 seconds on a single NVIDIA H20 GPU. Code, models, and datasets will be released upon paper acceptance.
\keywords{Image Super-Resolution \and One-Step 4K \and Periodic Artifact Suppression \and Flux}
\end{abstract}

\section{Introduction}
\label{sec:intro}

Single Image Super-Resolution (SISR) aims to recover original high-resolution (HR) images from degraded low-resolution (LR) images. Early approaches, such as CNN-based~\cite{srcnn,EDSR,nlsa,SAN}, Transformer-based~\cite{srformer,swinir,ATD,PFT}, Mamba-based~\cite{IHMambaSR,mambair,tamambair,mair}, and GAN-based methods~\cite{srgan,dasr,bsrgan,realesrgan} have made significant progress in SISR by leveraging local information, capturing long-range dependencies, modeling with global linear complexity, and utilizing adversarial learning. However, they still struggle to effectively generate complex texture details from severely degraded LR inputs.

Recently, diffusion-based SR~\cite{pocketsr,dong2025tinysr,zhu2025passionsr,dongjian1,dongjian2} has gained prominence for its superior data distribution modeling. Although leveraging Text-to-Image (T2I) priors yields remarkable perceptual quality compared to early train-from-scratch methods~\cite{ctmsr,resshift}, the heavy computational burden of iterative sampling has shifted recent focus toward accelerating the process to few- or even one-step inference.

\begin{figure*}[!t]
  \centering
  \includegraphics[width=1\linewidth]{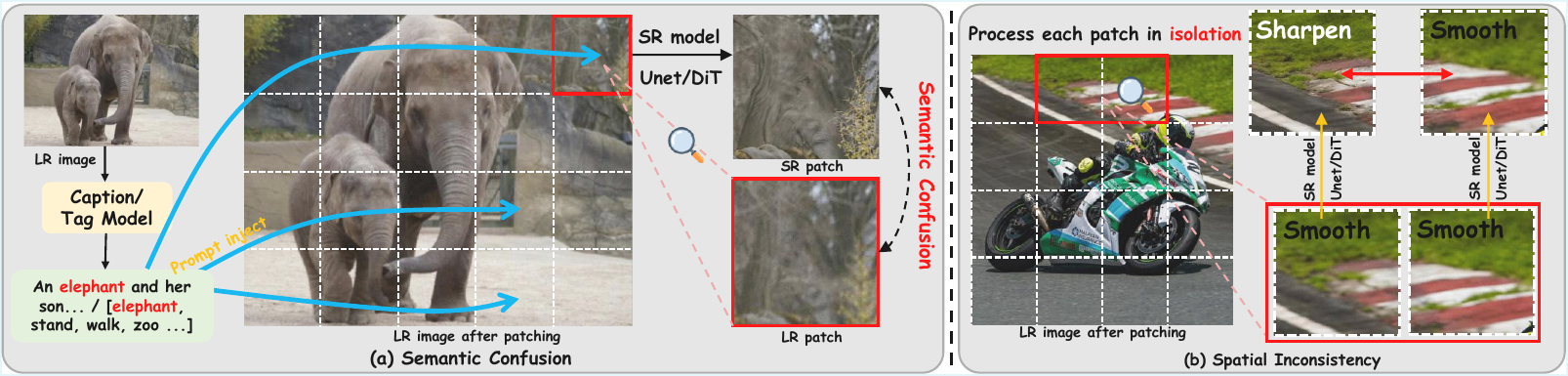}
\caption{Illustration of key limitations in patch-based 4K super-resolution. (a) Semantic confusion caused by prompt-dominated inference on local patches. (b) Spatial inconsistency arising from processing patches independently.}
  \vspace{-6.3mm}
  \label{fig:motivation1_2}
\end{figure*}

\newcommand{\cmark}{\textcolor{green!60!black}{\checkmark}}
\newcommand{\xmark}{\textcolor{red}{$\times$}}

\begin{wraptable}{r}{0.4\columnwidth} 
  \centering
  \vspace{-30pt} 
  \caption{Semantic consistency and spatial consistency support across different methods.}
  \label{tab:method_comparison}
  \resizebox{\linewidth}{!}{ 
  \begin{tabular}{lcc}
  \toprule
  \textbf{Method} & \textbf{\makecell{Semantic \\ Consistency}} & \textbf{\makecell{Spatial \\ Consistency}} \\
  \midrule
  ResShift~\cite{resshift}        & \cmark & \xmark \\
  SUPIR~\cite{supir }          & \xmark & \xmark \\
  DreamClear~\cite{ai2024dreamclear }      & \xmark & \xmark \\
  SinSR~\cite{sinsr}           & \cmark & \xmark \\
  AddSR~\cite{addsr}           & \xmark & \xmark \\
  OSEDiff~\cite{osediff}         & \xmark & \xmark \\
  OMGSR~\cite{omgsr}           & \cmark & \xmark \\
  \textbf{OP4KSR} & \cmark & \cmark \\
  \bottomrule
  \end{tabular}
  }
  \vspace{-15pt} 
\end{wraptable}

Driven by the rapid advancement of mobile devices and high-definition displays, the demand for 4K ultra-high-resolution images has surged. However, existing SR models utilizing T2I priors are primarily optimized for fixed, low-resolution domains (e.g., super-resolving images from $128\times128 \rightarrow 512\times 512$). Directly extrapolating these models to generate 4K images leads to severe generalization degradation and catastrophic memory bottlenecks. To circumvent this, current solutions widely adopt patch-based inference: partitioning the input, processing patches independently, and fusing the outputs. While computationally feasible, this spatial truncation introduces three critical flaws in 4K SR: \textbf{(i) Semantic confusion.} Isolated patches lacking global structural constraints are easily dominated by strong text priors, causing generative models to hallucinate mismatched semantics (e.g., injecting elephant textures into a background wall) (Fig.~\ref{fig:motivation1_2}(a)). \textbf{(ii) Spatial inconsistency.} Independent patch processing applies 
conflicting restoration strengths across continuous regions—aggressively 
sharpening one patch while over-smoothing an adjacent one—which yields 
abrupt visual transitions (Fig.~\ref{fig:motivation1_2}(b)). 
As further evidenced in Table~\ref{tab:method_comparison}, these two 
issues persist across existing SR methods. \textbf{(iii) Computational inefficiency.} Iterative patch processing incurs severe latency; specifically, super-resolving a $1024 \times 1024$ input to $4096 \times 4096$ using SUPIR~\cite{supir} takes $\sim$1600s on a single NVIDIA H20 GPU.

\begin{figure*}[!t]
  \centering
  \includegraphics[width=1\linewidth]{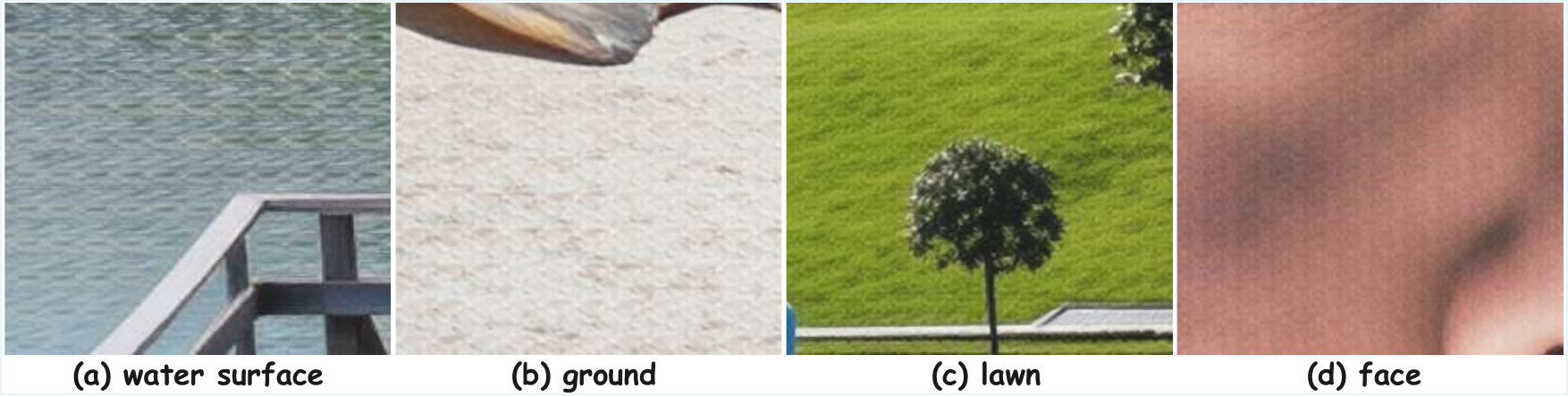}
\caption{Visual examples of periodic artifacts. These artifacts typically manifest in large, flat, and texturally homogeneous regions. We illustrate four representative scenes prone to such artifacts: (a) water surface, (b) ground, (c) lawn, and (d) face.}
  \vspace{-5mm}
  \label{fig:period——motivatoin}
\end{figure*}

\begin{wrapfigure}{r}{0.5\columnwidth}
  \centering
   \vspace{-1pt} 
  \includegraphics[width=\linewidth]{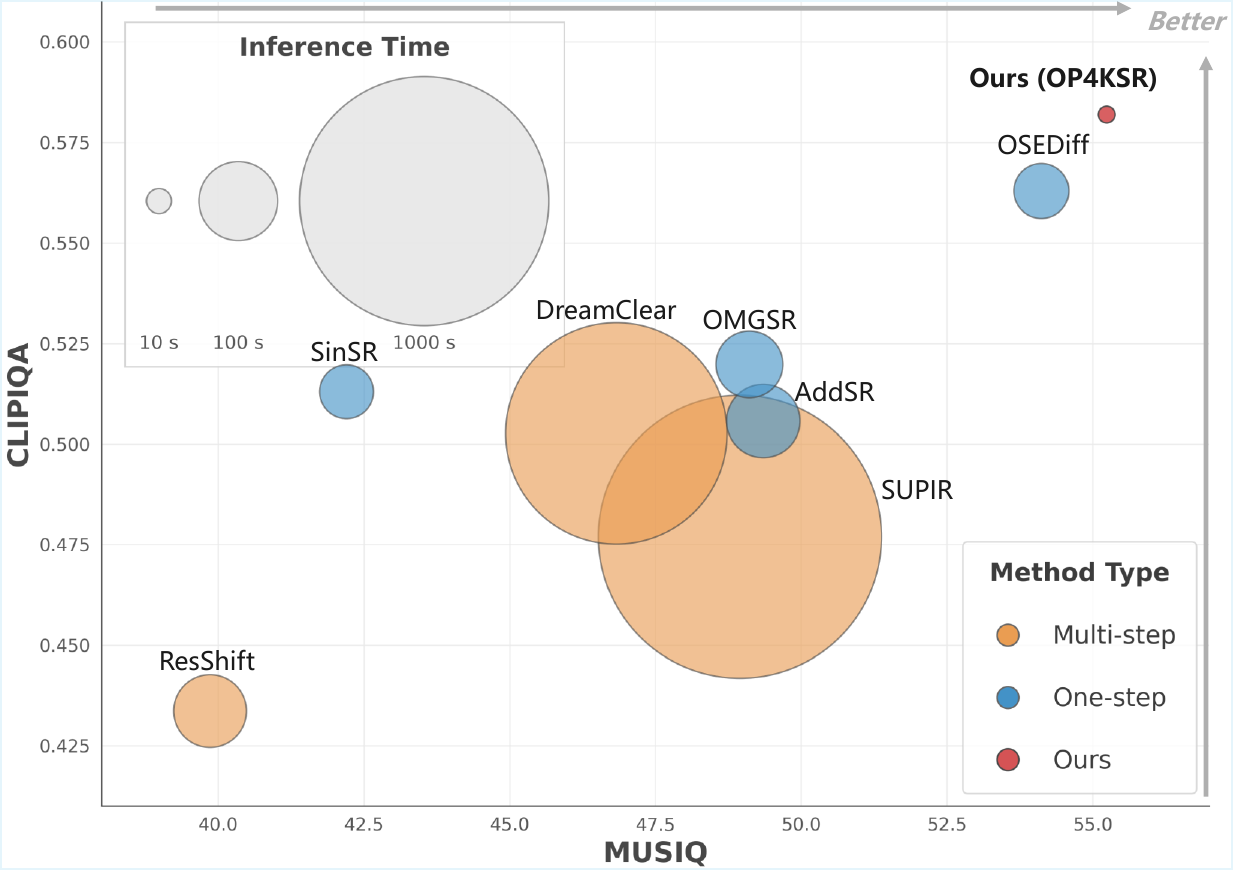}
\caption{Performance and runtime comparison of diffusion-based SR methods on the 4KSR-RealSquare benchmark ($1024 \times 1024 \rightarrow 4096 \times 4096$). OP4KSR achieves competitive reconstruction quality with significantly reduced inference time.}
  \label{fig:efficient_intro}
  \vspace{-20pt} 
\end{wrapfigure}

To overcome these limitations, we propose \textbf{OP4KSR}, a \textbf{o}ne-step, \textbf{p}atch-free framework for \textbf{4K} \textbf{SR} based on the Flux framework. First, we adopt an extreme-compression (F16) VAE to alleviate memory bottlenecks, unlocking global receptive fields for direct 4K SR. Second, we incorporate the mid-timestep strategy~\cite{omgsr} to  align latent features, enabling one-step SR. As illustrated in Fig.~\ref{fig:efficient_intro}, the framework empowers OP4KSR to achieve a nearly $\mathbf{10\times}$ speedup over one-step methods at 4K resolution while achieving competitive results.

However, adapting the multi-step pre-trained Flux framework to the one-step paradigm intrinsically triggers severe periodic artifacts (Fig.~\ref{fig:period——motivatoin}), 
which significantly degrade homogeneous areas like water surfaces and faces. We trace this flaw to two inherent causes: inter-token spatial indistinguishability arising from RoPE phase collapse, and intra-token spatial information loss during unpacking, both exacerbated by the lack of iterative refinement. Together, these compounding biases directly manifest as grid-like periodic artifacts. To resolve this, we introduce RoPE base frequency rescaling in the latent 
domain to compress ambiguous similarity zones and restore local structural sensitivity. In the pixel domain, a novel autocorrelation-based periodicity loss ($\mathcal{L}_{\text{AP}}$) locally constrains spatial autocorrelations, forcing the network to explicitly eliminate sub-grid periodicities rather 
than merely blurring them. Finally, we curate a high-quality 4K SR training dataset alongside three standardized benchmarks. In summary, our contributions are as follows:

\begin{itemize}
    \item We propose OP4KSR, to the best of our knowledge, the first one-step, patch-free framework for real-world 4K SR.
    \item We analyze the underlying causes of severe periodic artifacts in one-step Flux SR, and couple RoPE base frequency rescaling with a tailored autocorrelation-based periodicity loss to effectively mitigate these visual distortions.
    \item We construct a dedicated training set and three benchmarks (one synthetic, two real-world), effectively filling the critical gap in training resources and real-world benchmarks for 4K SR.
    \item Extensive experiments demonstrate that {OP4KSR} achieves highly competitive perceptual quality and the fastest inference speed among representative generative SR approaches.
\end{itemize}

\section{Related Works}

\textbf{Multi-step Diffusion-based Methods.}
Recently, benefiting from the strong generative priors encapsulated in large-scale pre-trained T2I models~\cite{sdxl,chen2024pixart,diffusion4k} (e.g., Stable Diffusion~\cite{blattmann2023stable} and Flux~\cite{flux}), diffusion models have demonstrated immense potential in SR~\cite{tvt,dit4sr,faithdiff,hypir}. DiffBIR~\cite{diffbir} employs a compact network to reconstruct LR images, subsequently applying ControlNet~\cite{controlnet} to guide the generation process. PASD~\cite{PASD} integrates pixel-level and semantic information via a pixel-aware cross-attention mechanism. SeeSR~\cite{seesr} introduces a dedicated module to extract semantic information from images. LucidFlux~\cite{lucidflux} utilizes a semantic alignment module to further extract fine-grained semantic details from images. However, these methods rely on multi-step inference, resulting in computational overhead that limits their deployment in real-world applications.

\noindent \textbf{One-step Diffusion-based Methods.}
One-step diffusion-based SR methods~\cite{adcsr,invsr,chen2025bridging,rcoe,tsdsr} have attracted increasing attention due to their significant practical potential. Specifically, SinSR~\cite{sinsr} effectively accelerates the ResShift~\cite{resshift} model through consistency-preserving distillation; OSEDiff~\cite{osediff} introduces Variational Score Distillation (VSD)~\cite{vsdloss} and directly takes the LR image as input, successfully compressing the multi-step denoising process into efficient one-step generation; AddSR~\cite{addsr} integrates Adversarial Diffusion Distillation (ADD) with the ControlNet architecture, simultaneously enabling efficient four-step and one-step inference models. Furthermore, PiSASR~\cite{pisasr} designs dual LoRA modules to optimize pixel-level and semantic-level objectives, thereby effectively balancing image fidelity and perceptual quality; OMGSR~\cite{omgsr} bridges the gap between real image inputs and the pre-trained noise latent distribution by devising a universal intermediate timestep training scheme and a latent space refinement loss.

\noindent \textbf{4K Image Super-Resolution.}
Early research on image SR focused on deterministic degradations (e.g., bicubic downsampling). Due to the simplicity of these degradations, the resulting models remained compact, making early exploration of 4K SR feasible. For instance, RT4KSR~\cite{rt4ksr} accelerates inference by downsampling deep features while preserving high-frequency details via a parallel branch. LRSRN~\cite{LRSRN} introduces reparameterized convolutions to enhance quality without extra overhead. CASR~\cite{casr} combines cascaded upsampling with knowledge distillation for ultra-fast 4K inference. Although these lightweight CNN-based methods achieve real-time 4K SR, they are inadequate for complex real-world scenarios. Meanwhile, most real-world SR approaches~\cite{stablesr,GTASR,upsr,zhu2024oftsr} are trained at low resolutions and still rely on region-wise inference when scaling to 4K, thereby inheriting the semantic and spatial limitations discussed in Section~\ref{sec:intro}.

\section{Method}

\subsection{One-Step 4K Super-Resolution Framework}

\begin{figure*}[!t]
  \centering
  \includegraphics[width=1\linewidth]{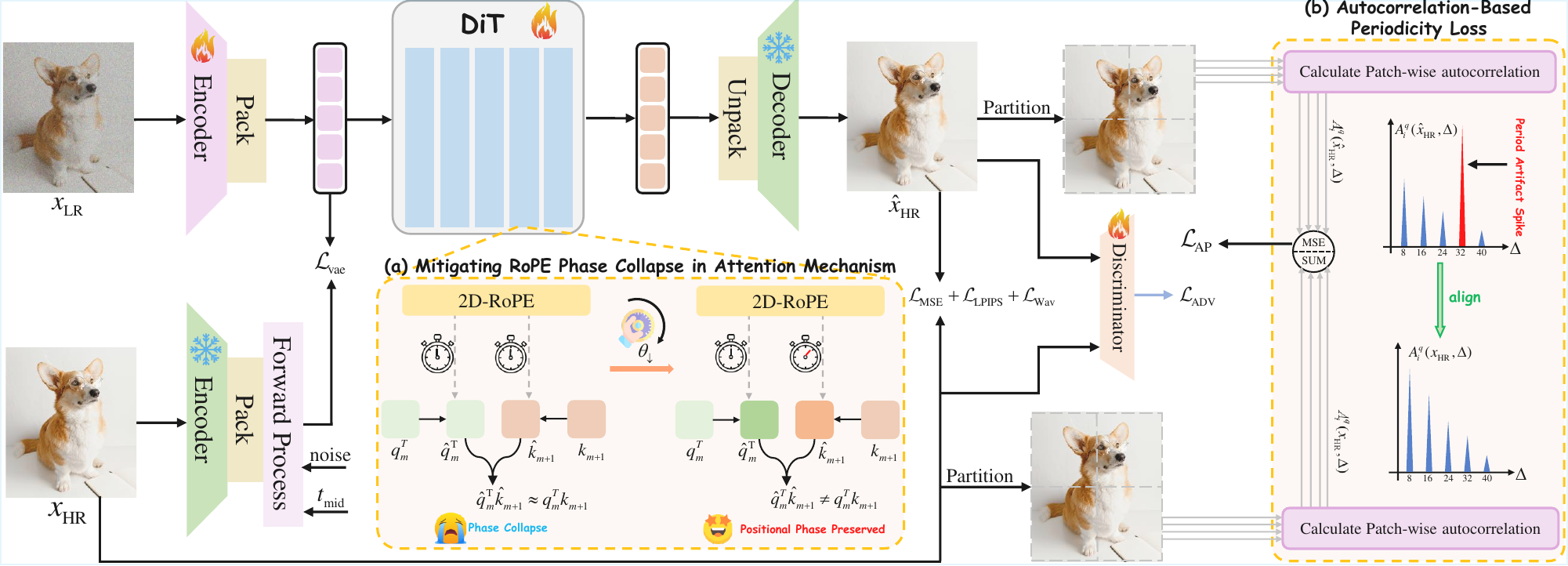}
\caption{ Overall architecture of OP4KSR. To suppress periodic artifacts, we introduce two components: (a) RoPE Base Frequency Rescaling (RFR), which lowers the RoPE base frequency $\theta$ to resolve inter-token ambiguity and restore fine-grained structural sensitivity; (b) Autocorrelation-based Periodicity Loss ($\mathcal{L}_{\text{AP}}$), which partitions images into $Q=4$ quadrants for independent autocorrelation computation to penalize periodic patterns. Detailed formulations of $\mathcal{L}_{\text{vae}}$~\cite{omgsr}, $\mathcal{L}_{\text{MSE}}$, $\mathcal{L}_{\text{LPIPS}}$, $\mathcal{L}_{\text{Wav}}$, and $\mathcal{L}_{\text{ADV}}$ are provided in the Supplementary Material.}
  \vspace{-5mm}
  \label{fig:overall}
\end{figure*}

To achieve patch-free, one-step 4K inference, OP4KSR builds upon UltraFlux~\cite{ultraflux}, utilizing its highly compressed F16 VAE encoder $E$ to bypass standard F8 constraints. Under the Flow Matching~\cite{lipman2022flow} framework, the intermediate latent $z_t$ linearly interpolates between pure noise $\epsilon$ and the HR latent $z_{\mathbf{HR}} = E(\mathbf{HR})$:
\begin{equation}
z_t = t z_{\mathbf{HR}} + (1-t)\epsilon.
\end{equation}
Adapting the pre-trained T2I framework for SR requires aligning the LR 
latent $z_{\mathbf{LR}} = E(\mathbf{LR})$ to an appropriate point on 
this flow. Naively injecting $z_{\mathbf{LR}}$ at the initial timestep 
creates a severe distribution gap, as the model expects pure noise 
rather than structured image content, disrupting its learned generative 
prior~\cite{fluxsr}. To this end, we adopt the latent alignment strategy 
from OMGSR~\cite{omgsr}. Specifically, we anchor $z_{\mathbf{LR}}$ at an optimal mid-timestep ($t_{\mathbf{mid}}=0.3$) to minimize the initial distribution gap, and further employ a latent alignment loss $\mathcal{L}_{\text{vae}}$ to explicitly constrain the intermediate latent $z_t$ to match $z_{\mathbf{LR}}$. This straightforward alignment seamlessly bridges the feature spaces, endowing OP4KSR with one-step SR capabilities. However, applying such one-step truncation to Flux-based 4K SR will introduce periodic artifacts.  To resolve this, we introduce a periodic artifact suppression strategy: a tailored RoPE base frequency rescaling (RFR) as shown in Fig.~\ref{fig:overall}(a) and an autocorrelation-based periodicity loss $\mathcal{L}_\text{AP}$ as shown in Fig.~\ref{fig:overall}(b). The overall architecture of OP4KSR is detailed in Fig.~\ref{fig:overall}.

\subsection{Overcoming Periodic Artifacts in One-Step Flux SR}
\label{sec:mitigating_artifacts}
\noindent\textbf{The Emergence of Periodic Artifacts.} Although integrating the UltraFlux prior enables highly efficient SR, we observe that the model's raw predictions suffer from catastrophic, grid-like artifacts with a precise fundamental period of 32 pixels---a spatial scale exactly matching one latent token (i.e., the $16\times$ VAE compression compounded by $2\times2$ token packing). These artifacts predominantly emerge in large, homogeneous regions or areas exhibiting highly repetitive textures as shown in Fig.~\ref{fig:period——motivatoin} (e.g., human faces, lawns, lake surfaces, and walls). We identify that this severe visual degradation stems from the convergence of two macroscopic shifts: (i) a drastic objective mismatch between T2I generation and SR, and  (ii) the aggressive step reduction in one-step inference. First, regarding the objective mismatch, pre-trained T2I models operate from pure noise, necessitating a strong focus on capturing long-range dependencies to synthesize overall structural contours. In contrast, the SR paradigm utilizes the LR image as a structural starting point, completely shifting the primary objective toward acquiring high-frequency local details. Second, regarding the inference efficiency, adapting UltraFlux for a single forward pass fundamentally strips away the multi-step iterative process, which typically acts as a crucial error-correction buffer. Without this buffer, the model fails to smooth out local token boundaries, causing latent spatial misalignments to manifest directly as visible grid artifacts.

\begin{figure*}[!t]
  \centering
    \vspace{-2mm}
  \includegraphics[width=1\linewidth]{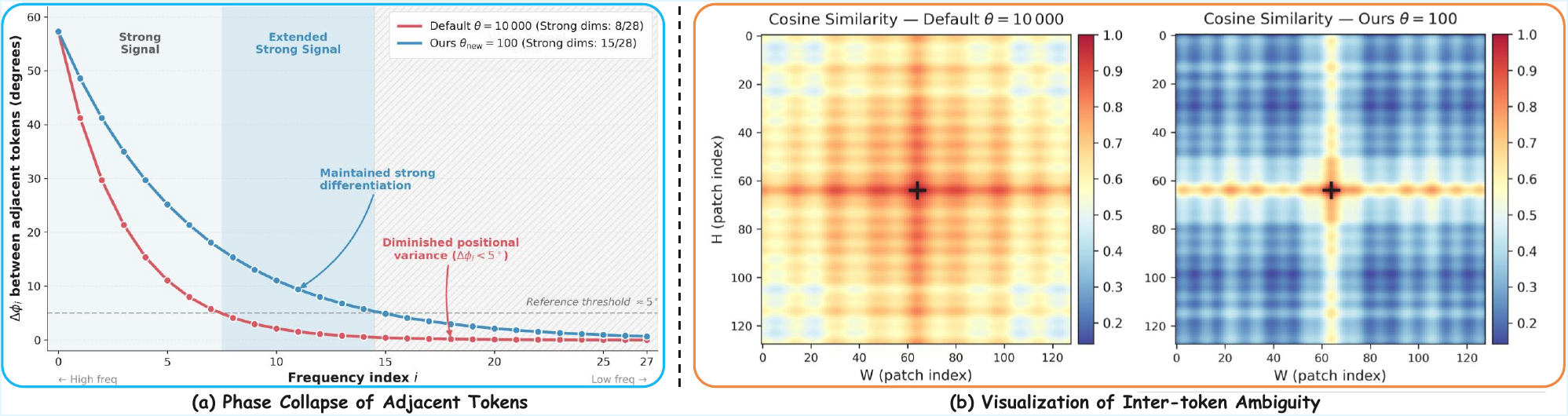}
\caption{Impact of RoPE base frequency $\theta$ on spatial perception. (a) Phase collapse of adjacent tokens: Under the default $\theta=10{,}000$, the relative phase difference $\Delta\phi_i$ rapidly collapses below $5^\circ$, leaving only 8/28 dimensions effective for local positioning. In contrast, our proposed $\theta=100$ extends the strong signal bandwidth to 15/28 dimensions. (b) Visualization of inter-token ambiguity: The default setting produces an extensive high-cosine-similarity region (red zone around the center query +), rendering adjacent tokens positionally indistinguishable. Our adjustment resolves this ambiguity and restores local distinguishability.}
  \vspace{-6mm}
  \label{fig:motivation2_rope}
\end{figure*}

\vspace{2mm}

\noindent\textbf{Diagnosing the Origins of Periodic Artifacts.} We attribute 
the observed 32-pixel grid artifacts to the interplay between the RoPE 
mechanism and the packed DiT architecture, which manifests at two distinct 
spatial granularities:

\textbf{(i)  RoPE Phase Collapse Leaves Adjacent Tokens Indistinguishable.} Originally introduced for 1D sequences in Large Language Models (LLMs), Rotary Positional Embedding (RoPE) is adapted into a 2D formulation in T2I models using a high base frequency (conventionally $\theta=10000$) to prioritize the aforementioned long-range dependencies. Directly transferring this global-centric configuration to the detail-oriented SR task introduces a severe frequency mismatch. Mathematically, following this 2D RoPE formulation, the rotation angle difference $\Delta\phi_i$ between spatially adjacent tokens is defined as:

\vspace{-1.8mm}

\begin{equation}
 \Delta\phi_i = \theta^{-\frac{2i}{d}}, \quad i \in \left\{0, 1, \dots, \frac{d}{2} - 1\right\},
\label{eq:rope_period}
\end{equation}
where $d$ is the per-axis feature dimension, $\theta$ is the base frequency, and $i$ represents the complex channel pair index. Under the $32\times$ compression ($16\times$ VAE and $2\times$pack), the latent spatial dimension for a 4096$\times$4096 image shrinks to  $128\times128$. 

Fundamentally, RoPE encodes positional information by applying a 2D rotation to feature pairs. For adjacent tokens (spatial distance $\Delta m = 1$), their spatial distinguishability is strictly governed by the relative rotation matrix $\mathbf{R}_{\Delta m=1}(\Delta\phi_i)$. However, with the default $\theta=10000$, the rotation angles $\Delta\phi_i$ for the vast majority of feature dimensions become pathologically small ($\Delta\phi_i < 5^\circ$) as shown in Fig.~\ref{fig:motivation2_rope}(a). Consequently, the rotational terms of the matrix critically approximate $\cos(\Delta\phi_i) \approx 1$ and $\sin(\Delta\phi_i) \approx 0$. Especially under BF16 precision, these minute rotational variations may
be further attenuated numerically, making $\mathbf{R}_{\Delta m=1}$
effectively close to the identity matrix $\mathbf{I}$. Therefore, when calculating the attention score between a RoPE-encoded 
query $\hat{\mathbf{q}}_m$ and an adjacent RoPE-encoded key 
$\hat{\mathbf{k}}_{m+1}$, their inner product collapses from 
$\hat{\mathbf{q}}_m^\top \hat{\mathbf{k}}_{m+1} = \mathbf{q}_m^\top 
\mathbf{R}_{\Delta m=1} \mathbf{k}_{m+1}$ directly to the unencoded 
base product $\mathbf{q}_m^\top \mathbf{k}_{m+1}$, completely dropping 
the positional phase shift. Without positional differentiation, 
attention scores across adjacent tokens converge to nearly identical 
values, producing the extensive high-similarity region in 
Fig.~\ref{fig:motivation2_rope}(b). This leaves 
the network fundamentally unable to resolve local spatial relationships. 
Further details are provided in the Supplementary Material.

\begin{wrapfigure}{r}{0.6\columnwidth}
  \centering
  \vspace{-20pt} 
  \includegraphics[width=\linewidth]{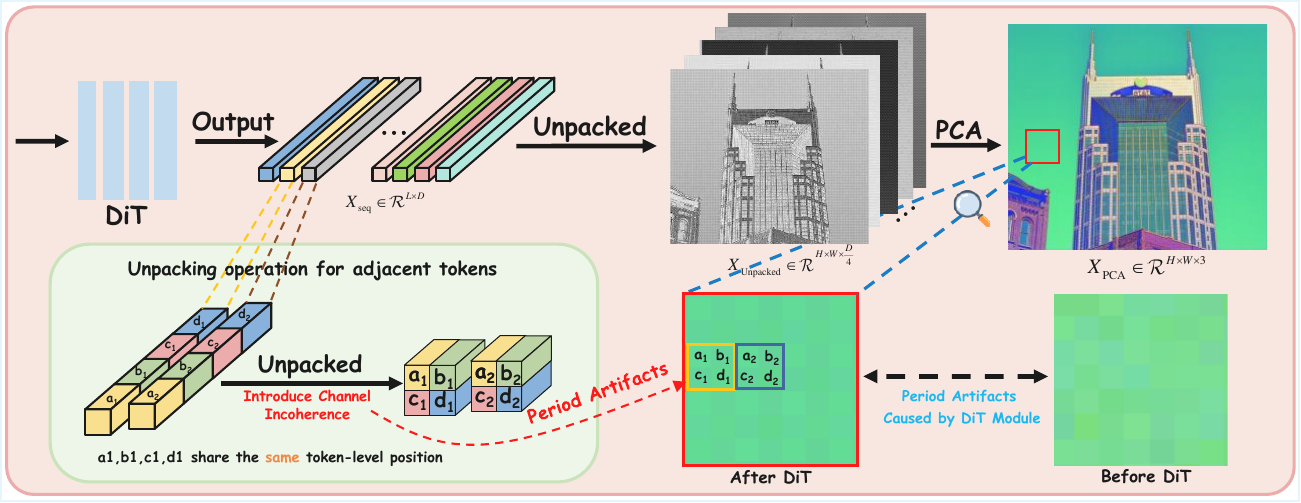}
\caption{Illustration of intra-token blindness. Left: The \texttt{Unpack} operation rearranges channel-wise sub-pixel features ($a_1, b_1, c_1, d_1$) into a $2 \times 2$ spatial grid. Right: PCA visualization reveals periodic grid artifacts in $X_{\text{Unpacked}}$ after DiT, in contrast to the artifact-free latents before DiT. This confirms that the periodic artifacts originate from the DiT module.}
  \label{fig:unpack}
  \vspace{-20pt} 
\end{wrapfigure}

\textbf{(ii) Unpacking Channels Lacking Spatial Information Results in Periodic Artifacts.} Compounding the inter-token blindness caused by the RoPE collapse is a secondary geometric collapse within the latent token itself. As illustrated in Fig.~\ref{fig:unpack}, inside the DiT, the features corresponding to a $2 \times 2$ spatial patch (sub-pixels $a_1, b_1, c_1, d_1$) reside entirely within the channel dimension of a single latent token. Because the standard DiT architecture injects positional encoding strictly at the token level, these sub-pixel features are forced to share the exact same position throughout the transformer layers. While these features are separated into distinct channels, channel dimensions inherently lack spatial semantics. Consequently, the network cannot differentiate their internal geometry (e.g., distinguishing the top-left $a_1$ from the bottom-right $d_1$). This critical vulnerability ultimately materializes during the \texttt{Unpack} step: as the network deterministically unfolds these positionally agnostic channels back into a spatial grid, it intrinsically introduces severe structural misalignments.

 Crucially, whether this intra-token incoherence manifests as macroscopic periodic artifacts depends entirely on the local image content. In flat, homogeneous regions, unencoded adjacent features are inherently similar. Since the phase collapse ($\mathbf{R}_{\Delta m} \to \mathbf{I}$, analyzed in \textbf{(i)}) prevents RoPE from breaking this symmetry, the network is forced to aggregate indistinguishable tokens into uniform latent clumps (i.e., $a_1 \approx a_2, b_1 \approx b_2$). Consequently, the structurally biased pattern ($a_1, b_1, c_1, d_1$) produced by the \texttt{Unpack} operation is mechanically repeated across adjacent regions, deterministically magnifying into catastrophic periodic grid artifacts. Conversely, in high-frequency regions, adjacent base features inherently possess substantial variance. Even without effective positional encoding, this natural discrepancy organically breaks the structural symmetry ($a_1 \neq a_2$), disrupting the repetitive tiling process and explaining why these artifacts predominantly afflict smooth areas.

This mechanism is directly corroborated by our PCA visualization in Fig.~\ref{fig:unpack}: while the latent features before the DiT remain structurally smooth, the $X_{\text{Unpacked}}$ latents after the DiT exhibit the $a_1, b_1, c_1, d_1$ periodic grid. This confirms that the fatal synergy between token-level PE sharing and the \texttt{Unpacked} operation is the direct structural origin of the periodic artifacts upon VAE decoding.

\vspace{2mm}

\noindent \textbf{Periodic Artifact Suppression Strategy.} To address the RoPE phase collapse and the spatial information loss in unpacking channels, we propose a synergistic strategy operating in both the latent and pixel domains:

\textbf{(i) RoPE Base Frequency Rescaling.} To resolve the inter-token ambiguity identified in our prior analysis, we propose RoPE base frequency rescaling (RFR), which significantly reduces the base frequency $\theta$ from $10000$ to $100$. As quantitatively demonstrated in Fig.~\ref{fig:motivation2_rope}(a), this adjustment decreases the decay rate of the rotation angle differences $\Delta\phi_i$. By aligning the frequency spectrum with the compact latent resolution, RFR extends the effective bandwidth of strong positional signals ($\Delta\phi_i > 5^\circ$) from 8 to 15 channel dimensions. The effectiveness of this rescaling is visually confirmed in Fig.~\ref{fig:motivation2_rope}(b): the ambiguous high-similarity zone produced by the default setting is substantially compressed. While a large base frequency $\theta$ blurs the distinction between adjacent tokens, essentially rendering the positional injection ineffective, our $\theta=100$ configuration explicitly restores the model's ability to discriminate fine-grained spatial 
relations among neighboring tokens, effectively breaking the latent 
spatial ambiguity.

\textbf{(ii) Autocorrelation-based Periodicity Loss.} To compensate for the intra-token spatial ambiguity intrinsic to one-step generation, we introduce an autocorrelation-based periodicity loss ($\mathcal{L}_{\text{AP}}$) to explicitly suppress the 32-pixel grid artifacts (Fig.~\ref{fig:overall}(b)). As previously observed, these artifacts primarily affect localized smooth regions. A global autocorrelation computation~\cite{dip} would inevitably dilute these concentrated periodic signals with the natural frequencies of artifact-free areas. To prevent this spatial averaging effect and accurately isolate the anomalies, we partition both the predicted HR image $\hat{{x}}_{\text{HR}}$ and its ground-truth ${x}_{\text{HR}}$ into $Q=4$ distinct quadrants, computing the unbiased 2D spatial autocorrelation $A(x,\Delta)$ independently for each block $x$ at spatial lag $\Delta$. Furthermore, to avoid naive artifact blurring and localized over-smoothing, it is insufficient to penalize solely the fundamental 32-pixel peak. Instead, we constrain the autocorrelation alignment over a comprehensive set of evenly spaced spatial lags: $K = \{8, 16, 24, 32, 40\}$. By matching the autocorrelation responses across these multiple intervals, we compel the network to faithfully reproduce the intrinsic spatial correlation properties of the ground truth, rather than merely suppressing the anomalies. The final objective minimizes the Mean Squared Error (MSE) between the evaluated autocorrelation responses $A(\hat{x}, \Delta)$ and $A(x, \Delta)$ across all partitioned blocks and targeted lags:

\vspace{-6mm}

\begin{equation}{
\mathcal{L}_{\text{AP}} = \frac{1}{2 Q C |K|} \sum_{q=1}^{Q} \sum_{c=1}^{C} \sum_{\Delta \in K} \sum_{i \in \{\text{h}, \text{v}\}} \left( A_{i,c}^q(\hat{{x}}_{\text{HR}}, \Delta) - A_{i,c}^q({x}_{\text{HR}}, \Delta) \right)^2,}
\vspace{-1mm}
\end{equation}
where $C$ denotes the image channels, $i \in \{\text{h}, \text{v}\}$ indicates the horizontal and vertical spatial axes, and ${x}_{\text{HR}}$ is explicitly detached from the computation graph. 
The specific details of $\mathcal{L}_{\text{AP}}$ are provided in the 
Supplementary Material.

\subsection{4K Dataset Construction}

\begin{wrapfigure}{r}{0.45\columnwidth}
  \centering
  \vspace{-30pt} 
  \includegraphics[width=\linewidth]{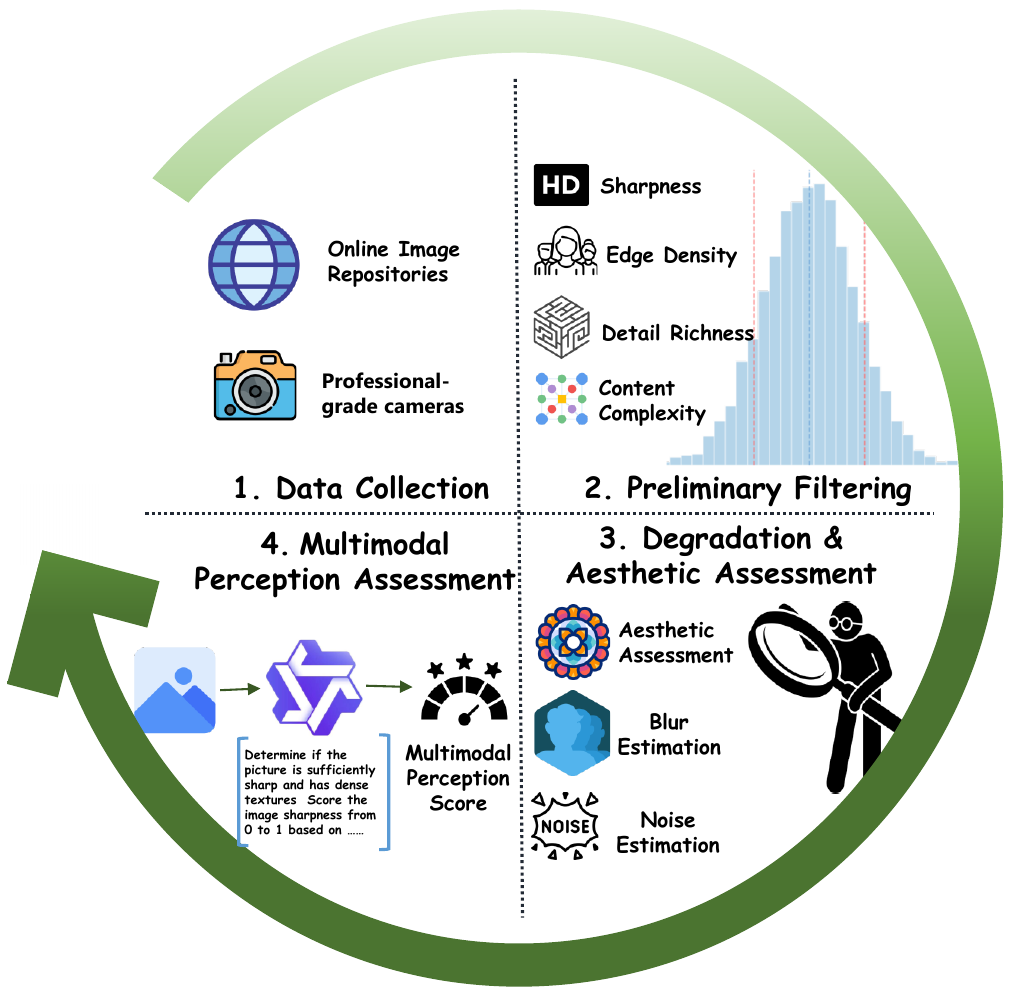}
\caption{Data curation pipeline for the 4KSR-Train dataset. Raw images are collected at scale and processed through a three-stage filtering pipeline: (1) Preliminary Filtering, (2) Degradation \& Aesthetic Assessment, and (3) Multimodal Quality Assessment.}
  \label{fig:datapipeline}
  \vspace{-20pt} 
\end{wrapfigure}

\vspace{2mm}

Existing SR datasets (e.g., DIV2K~\cite{div}, LSDIR~\cite{lsdir} and Flickr2K~\cite{flickr}) are limited in resolution, while publicly available high-fidelity 4K resources remain scarce. This scarcity severely hinders the development and accurate evaluation of ultra-high-resolution models. To effectively fill this critical gap in both training resources and real-world evaluation benchmarks, we curate a dedicated 4K SR suite (summarized in Table~\ref{tab:4ksr_datasets}), comprising a training dataset, {4KSR-Train}, alongside three comprehensive evaluation benchmarks: the synthetic benchmark {4KSR-Syn} and two real-world benchmarks, {4KSR-RealSquare} and {4KSR-RealVary}.

\noindent\textbf{4KSR-Train Dataset.} 
We initially curate a corpus of $\sim$1.1M high-quality images from the internet and real-world photography. To ensure high fidelity and texture richness, we design a rigorous three-stage filtering pipeline (depicted in Fig.~\ref{fig:datapipeline}):

\textbf{(i) Preliminary Filtering:} We apply multiple complementary filters to ensure basic visual quality 
and complexity. First, we employ Laplacian variance and the Sobel operator to eliminate blurred and overly flat images. Next, to guarantee detail richness and content complexity, we compute Gray-Level Co-occurrence Matrix (GLCM) metrics (capturing contrast and correlation) and Shannon entropy. Only the top 50\% of images based on aggregated GLCM scores and entropy are retained.

\textbf{(ii) Degradation \& Aesthetic Assessment:} To guarantee both technical fidelity and perceptual realism, we employ a dual-criterion filtering strategy. First, we adapt DEResNet~\cite{mou2022metric} to quantify degradation levels, discarding samples with excessive blur and noise artifacts. Second, to align with human visual preferences, we utilize the LAION Aesthetic Predictor~\cite{schuhmann2022laion} to evaluate visual composition and color harmony.

\textbf{(iii) Multimodal Quality Assessment:} Finally, we leverage Qwen3-VL~\cite{bai2025qwen3} to conduct a comprehensive evaluation encompassing texture richness and clarity. This step ensures the retention of samples exhibiting high semantic integrity and superior perceptual fidelity.

\vspace{2mm}

\noindent\textbf{Evaluation Benchmarks.}
We constructed three benchmarks to comprehensively evaluate 4K SR performance.

\textbf{(i) Synthetic Benchmark (4KSR-Syn):} We reserved 82 diverse images from the source pool (strictly excluded from training) and cropped such that the maximum side length of each HR image is 4096.
The corresponding LR inputs were generated using a recalibrated Real-ESRGAN~\cite{realesrgan} pipeline tailored for 4K.

\textbf{(ii) Real-World Benchmarks (4KSR-Real):} To assess robustness against authentic degradations, we collected two unpaired real-world datasets: \textbf{4KSR-RealSquare} (92 isotropic $1024 \times 1024$ images) and \textbf{4KSR-RealVary} (81 images with diverse aspect ratios). Details are provided in the Supplementary Material.


\begin{table}[!t]
    \centering
    \caption{\textbf{Overview of the proposed 4KSR dataset suite.}}
    \label{tab:4ksr_datasets}
    \setlength{\tabcolsep}{10pt}
    \vspace{-1mm}
    
    \resizebox{0.92\linewidth}{!}{%
    \begin{tabular}{l c c c c c c}
        \toprule
        \textbf{Dataset} & \textbf{Role} & \textbf{Source} & \textbf{Resolution} & \textbf{Count} & \textbf{GT} & \textbf{Characteristics} \\
        \midrule
        \textbf{4KSR-Train} & Train & Hybrid & Mixed ($<4$K) & 34,379 & \cmark & High-fidelity, Texture-rich \\
        \midrule
        \textbf{4KSR-Syn} & Test & Synthetic & Max edge 1024 & 82 & \cmark & Strict 4K, Full-reference \\
        \textbf{4KSR-RealSquare} & Test & Real-world & $1024 \times 1024$ & 92 & \xmark & Isotropic, Authentic noise \\
        \textbf{4KSR-RealVary} & Test & Real-world & Max edge 1024 & 81 & \xmark & Varied Aspect Ratios \\
        \bottomrule
    \end{tabular}%
    }
    
    \vspace{-4mm}
\end{table}

 \section{Experimental Results}

\subsection{Experimental settings}

\noindent  \textbf{Training Datasets and Benchmarks.}
We train our OP4KSR on our proposed {4KSR-Train} dataset. For evaluation, alongside our rigorously constructed 4KSR suite ({4KSR-Syn}, {4KSR-RealSquare}, and {4KSR-RealVary}), we incorporate the public dataset {DIV4K-50}~\cite{zuo20254kagent} to assess synthetic generalization. For fair comparison, its LR inputs are generated using our training degradation pipeline.

\begin{table*}[t]
\centering
\footnotesize
\setlength{\tabcolsep}{2pt} 
\renewcommand{\arraystretch}{1}

\caption{Quantitative comparison on synthetic benchmarks (4KSR-Syn and DIV4K-50). The best and second-best results are highlighted in \textbf{bold} and \underline{underline}.}
\label{tab:synthetic_results}

\scalebox{0.72}{
\begin{tabular}{cl ccc ccccc} 
\toprule
\multirow{2}{*}{Datasets} & \multirow{2}{*}{Metrics} 
& \multicolumn{3}{c}{Diffusion-based multi-step methods} 
& \multicolumn{5}{c}{Diffusion-based one-step methods} \\
\cmidrule(lr){3-5} \cmidrule(lr){6-10} 
& 
& ResShift & SUPIR & DreamClear
& SinSR & AddSR & OSEDiff & OMGSR & \textbf{OP4KSR} \\

\multicolumn{2}{c}{{T2I Prior}} & -- & SDXL & PixArt-$\alpha$ & -- &SD 2-base &SD 2.1-base & Flux.1 & UltraFlux \\
\midrule

\multirow{12}{*}{\textbf{4KSR-Syn}}
& PSNR$\uparrow$       & \textbf{27.70}  & 26.31  & 24.32  & \underline{27.06}  & 26.45  & 25.79  & 24.62  & 24.31  \\
& SSIM$\uparrow$       & \textbf{0.7459} & 0.6884 & 0.6580 & 0.6990 & \underline{0.7342} & 0.7200 & 0.6218 & 0.6830 \\
& LPIPS$\downarrow$    & 0.4028 & 0.3571 & 0.3781 & 0.4208 & 0.3523 & \underline{0.3478} & 0.4515 & \textbf{0.3297} \\
& DISTS$\downarrow$    & 0.2055 & \underline{0.1533} & 0.1617 & 0.1933 & 0.1858 & 0.1678 & 0.1935 & \textbf{0.1496} \\
& TOPIQ\_FR$\uparrow$  & 0.4311 & 0.4445 & 0.4059 & 0.4459 & 0.4007 & \underline{0.4472} & 0.3958 & \textbf{0.4507} \\
& NIQE$\downarrow$     & 5.82   & \underline{3.12}   & \textbf{2.84}   & 5.57   & 4.78   & 3.77   & 5.38   & 4.80   \\
& MANIQA$\uparrow$     & 0.3475 & 0.3840 & \underline{0.4103} & 0.3734 & 0.4050 & \textbf{0.4587} & 0.3508 & 0.3931 \\
& MUSIQ$\uparrow$      & 42.13  & 52.52  & 51.70  & 45.26  & 48.62  & \underline{55.10}  & 49.20  & \textbf{56.51}  \\
& CLIPIQA$\uparrow$    & 0.5275 & 0.5776 & 0.5994 & 0.5925 & 0.5718 & \textbf{0.6578} & 0.5238 & \underline{0.6135} \\
& TOPIQ\_NR$\uparrow$  & 0.4238   & 0.5034   & 0.5025   & 0.4694   & {0.5102}   & \textbf{0.5803}   & 0.4104   & \underline{0.5159}   \\
& TOPIQ\_IAA$\uparrow$ & 4.19   & 4.33   & 4.31   & 4.28   & 4.20   & \textbf{4.43}   & 4.16   & \underline{4.37}   \\
& NIMA$\uparrow$       & 4.41   & 4.51   & 4.51   & 4.48   & 4.28   & \underline{4.62}   & 4.52   & \textbf{4.68}   \\
\midrule 

\multirow{12}{*}{\textbf{DIV4K-50}}
& PSNR$\uparrow$       & \textbf{26.21}  & 24.73  & 23.33  & \underline{25.33}  & 24.85  & 24.72  & 23.80  & 23.09  \\
& SSIM$\uparrow$       & \textbf{0.6304} & 0.5694 & 0.5554 & 0.5692 & \underline{0.6191} & 0.6090 & 0.5292 & 0.5616 \\
& LPIPS$\downarrow$    & 0.5506 & \underline{0.4640} & 0.4744 & 0.5523 & 0.5193 & 0.4769 & 0.5389 & \textbf{0.4461} \\
& DISTS$\downarrow$    & 0.2562 & \underline{0.1924} & 0.2057 & 0.2324 & 0.2420 & 0.2161 & 0.2329 & \textbf{0.1843} \\
& TOPIQ\_FR$\uparrow$  & 0.3982 & 0.4131 & 0.3747 & \underline{0.4159} & 0.3575 & 0.4096 & 0.3829 & \textbf{0.4263} \\
& NIQE$\downarrow$     & 6.62   & \underline{3.14}   & \textbf{2.88}   & 6.20   & 4.64   & 3.93   & 5.54   & 5.04   \\
& MANIQA$\uparrow$     & 0.2643 & 0.3228 & 0.3648 & 0.3051 & \underline{0.3740} & \textbf{0.4056} & 0.3077 & 0.3361 \\
& MUSIQ$\uparrow$      & 33.36  & 46.81  & 47.46  & 38.03  & 47.71  & \underline{50.11}  & 42.68  & \textbf{52.24}  \\
& CLIPIQA$\uparrow$    & 0.4345 & 0.5145 & 0.5379 & 0.5039 & 0.5267 & \textbf{0.6044} & 0.4630 & \underline{0.5409} \\
& TOPIQ\_NR$\uparrow$  & 0.3464   & 0.4593   & \underline{0.5321}   & 0.4233   & 0.4987   & \textbf{0.5401}   & 0.3724   & 0.4703   \\
& TOPIQ\_IAA$\uparrow$ & 4.11   & 4.44   & 4.32   & 4.27   & \underline{4.44}   & \textbf{4.57}   & 4.14   & 4.29   \\
& NIMA$\uparrow$       & 4.25   & 4.51   & 4.35   & 4.35   & 4.47   & \underline{4.48}   & 4.40   & \textbf{4.55}   \\
\bottomrule
\vspace{-10mm}
\end{tabular}}
\end{table*}

\noindent  \textbf{Evaluation metrics.} We evaluate all methods using full-reference and no-reference image quality metrics. For full-reference assessment, we employ PSNR and SSIM~\cite{wang2004image} computed on the Y channel in YCbCr space to evaluate reconstruction fidelity, alongside LPIPS~\cite{zhang2018unreasonable}, DISTS~\cite{distst}, and TOPIQ-FR~\cite{chen2024topiq} to measure perceptual similarity. The no-reference metrics include NIQE~\cite{zhang2015feature}, 
MANIQA~\cite{yang2022maniqa}, MUSIQ~\cite{ke2021musiq}, CLIPIQA~\cite{wang2023exploring},  TOPIQ-NR~\cite{chen2024topiq}, TOPIQ-IAA~\cite{chen2024topiq}, and NIMA~\cite{talebi2018nima}, which estimate perceptual quality without references. Unlike traditional fidelity and statistical metrics (e.g., PSNR and NIQE), deep-feature-based metrics (e.g., CLIPIQA and LPIPS) are inherently resolution-sensitive. Thus, we adopt a patch-based evaluation strategy, as detailed in the Supplementary Material.

\begin{table*}[t]
\centering
\footnotesize
\setlength{\tabcolsep}{2pt} 
\renewcommand{\arraystretch}{1}

\caption{Quantitative comparison on real-world benchmarks (4KSR-RealSquare and 4KSR-RealVary). The best and second-best are highlighted in \textbf{bold} and \underline{underline}. }
\label{tab:realworld_results}

\scalebox{0.72}{
\begin{tabular}{cl ccc ccccc} 
\toprule
\multirow{2}{*}{Datasets} & \multirow{2}{*}{Metrics} 
& \multicolumn{3}{c}{Diffusion-based multi-step methods} 
& \multicolumn{5}{c}{Diffusion-based one-step methods} \\
\cmidrule(lr){3-5} \cmidrule(lr){6-10} 
& 
& ResShift & SUPIR & DreamClear
& SinSR & AddSR & OSEDiff & OMGSR & \textbf{OP4KSR} \\

\multicolumn{2}{c}{{T2I Prior}} & -- & SDXL & PixArt-$\alpha$ & -- & SD 2-base & SD 2.1-base & Flux.1 & UltraFlux \\
\midrule

\multirow{7}{*}{\begin{tabular}{@{}c@{}}\textbf{4KSR-} \\ \textbf{RealSquare}\end{tabular}}
& NIQE$\downarrow$     & 6.26   & \underline{3.20}   & \textbf{3.16}   & 5.85   & 4.92   & 3.89   & 5.31   & 4.93   \\
& MANIQA$\uparrow$     & 0.2864 & 0.3313 & 0.3460 & 0.3213 & \underline{0.3732} & \textbf{0.4213} & 0.3296 & 0.3489 \\
& MUSIQ$\uparrow$      & 39.86  & 48.95  & 46.83  & 42.20  & 49.35  & \underline{54.12}  & 49.11  & \textbf{55.24}  \\
& CLIPIQA$\uparrow$    & 0.4337 & 0.4770 & 0.5027 & 0.5130 & 0.5058 & \underline{0.5630} & 0.5198 & \textbf{0.5820} \\
& TOPIQ\_NR$\uparrow$  & 0.3729   & 0.4501   & 0.4858   & 0.4121   & 0.4879   & \textbf{0.5322}   & 0.4029   & \underline{0.5034}   \\
& TOPIQ\_IAA$\uparrow$ & 4.04   & 4.05   & 3.93   & \underline{4.13}   & 4.08   & 4.06   & 3.97   & \textbf{4.15}   \\
& NIMA$\uparrow$       & 4.28   & 4.24   & 4.24   & 4.29   & 4.17   & \underline{4.42}   & 4.40   & \textbf{4.47}   \\
\midrule 

\multirow{7}{*}{\begin{tabular}{@{}c@{}}\textbf{4KSR-} \\ \textbf{RealVary}\end{tabular}}
& NIQE$\downarrow$     & 6.27   & \underline{3.38}   & \textbf{3.27}   & 5.90   & 4.83   & 3.74   & 5.24   & 4.92   \\
& MANIQA$\uparrow$     & 0.3095 & 0.3253 & 0.3669 & 0.3438 & \underline{0.3859} & \textbf{0.4324} & 0.3379 & 0.3617 \\
& MUSIQ$\uparrow$      & 42.92  & 49.14  & 49.46  & 45.43  & 51.26  & \underline{56.02}  & 51.19  & \textbf{57.13}  \\
& CLIPIQA$\uparrow$    & 0.4660 & 0.4581 & 0.5040 & 0.5485 & 0.5176 & \underline{0.5679} & 0.5191 & \textbf{0.5831} \\
& TOPIQ\_NR$\uparrow$  & 0.4094   & 0.4432   & 0.4799   & 0.4416   & \textbf{0.5234}   & \underline{0.5223}   & 0.4139   & 0.5068   \\
& TOPIQ\_IAA$\uparrow$ & \underline{4.15}   & 4.06   & 3.96   & 4.13   & 4.12   & 4.13   & 4.00   & \textbf{4.20}   \\
& NIMA$\uparrow$       & 4.36   & 4.20   & 4.28   & 4.35   & 4.21   & 4.39   & \underline{4.41}   & \textbf{4.46}   \\
\bottomrule
 \vspace{-3mm}
\end{tabular}}
\end{table*}

\begin{figure*}[!t]
  \centering
  \includegraphics[width=1\linewidth]{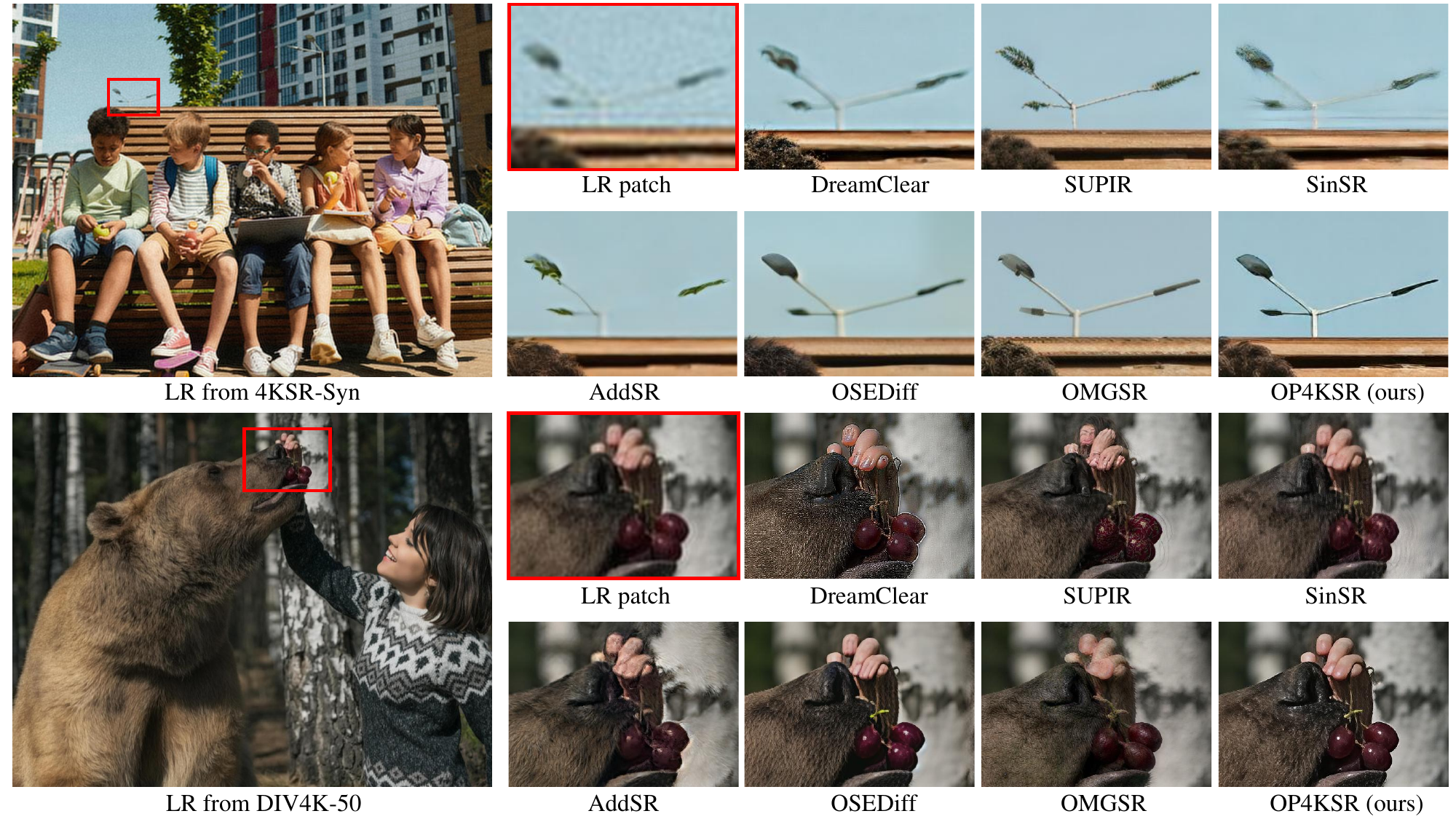}
  \caption{Qualitative comparisons of different methods on  two \textbf{synthetic datasets.}}
  \vspace{-7mm}
  \label{fig:syn}
\end{figure*}

\begin{figure*}[!t]
  \centering
  \includegraphics[width=1\linewidth]{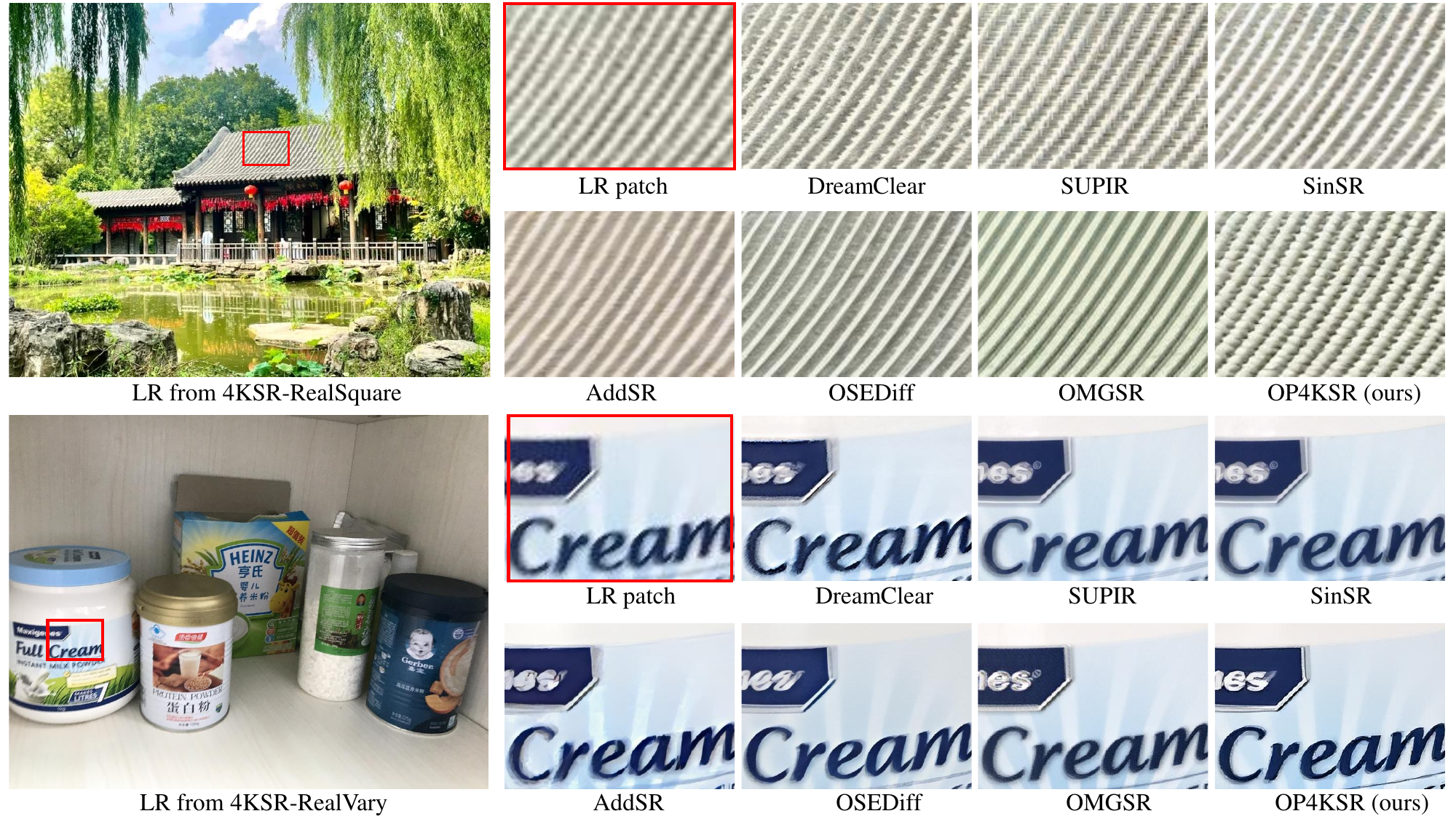}
\caption{Qualitative comparisons of different methods on two \textbf{real-world datasets.}  }
  \vspace{-6mm}
  \label{fig:real}
\end{figure*}

\noindent  \textbf{Implementation Details.}
Following~\cite{osediff,omgsr}, we freeze the VAE decoder and exclusively fine-tune the VAE encoder and the DiT backbone using LoRA, with the LoRA rank set to 64 for both. To reduce the computational overhead associated with text feature extraction, we adopt fixed text prompts within the S3Diff pipeline~\cite{s3diff}. During the training phase, we utilize the AdamW optimizer with a learning rate of $2.5\times10^{-5}$. The experiments are conducted on 8 H20 GPUs with an effective batch size of 16. The entire training process spans 2,000 iterations and takes approximately 2 days. 

\subsection{Comparison with the state of the art}

\textbf{Compared methods.} We compare our proposed OP4KSR with several representative approaches, including diffusion-based multi-step methods (ResShift~\cite{resshift}, SUPIR~\cite{supir}, and DreamClear~\cite{ai2024dreamclear}) and diffusion-based one-step methods (SinSR~\cite{sinsr}, AddSR~\cite{addsr}, OSEDiff~\cite{osediff} ,and OMGSR~\cite{omgsr}). Specifically, we evaluate the 15-step version of ResShift and the Flux-based variant of OMGSR.

\noindent \textbf{Quantitative Comparisons.}
 On the synthetic datasets reported in Table~\ref{tab:synthetic_results}, our method achieves the best perceptual fidelity, obtaining the lowest LPIPS and DISTS as well as the highest TOPIQ-FR among all compared approaches. In addition, OP4KSR delivers competitive aesthetic quality with leading MUSIQ and NIMA scores, indicating its superior capability in generating visually pleasing and authentic details. Results on the real-world datasets in Table~\ref{tab:realworld_results} further validate its robustness. OP4KSR achieves highly competitive performance in no-reference evaluations, delivering strong perceptual quality over other methods.

\begin{table}[!t]
  \centering
   \setlength{\tabcolsep}{10pt}
  \caption{Comparison of inference time and memory usage across different methods.}
  \label{tab:efficiency}
  \resizebox{0.85\linewidth}{!}{
    \begin{tabular}{l c c c c c c c c}
      \toprule[1.5pt] 
      
      & \multicolumn{3}{c}{Diffusion-based full-step methods} 
      & \multicolumn{5}{c}{Diffusion-based one-step methods} \\
      
      \cmidrule(lr){2-4} \cmidrule(lr){5-9} 
      
      Metrics 
      & ResShift & SUPIR & DreamClear 
      & SinSR & AddSR & OSEDiff & OMGSR & \textbf{OP4KSR} \\
      
      {{T2I Prior}} 
      & -- & SDXL & PixArt-$\alpha$ 
      & -- & SD 2-base & SD 2.1-base & Flux.1 & UltraFlux \\
      \midrule
          
      Inference time (s) $\downarrow$   
      & 108.8 & 1643.7 & 1006.5 
      & \underline{59.6} & 110.1 & 61.77 & 91.33 & \textbf{5.75} \\
      
      Memory usage (G) $\downarrow$ 
      & 36.81 & 53.26  & 64.13
      & \textbf{25.88} & 37.71 & 46.53 & 64.10 & \underline{32.88} \\
      
      \bottomrule[1.5pt] 
    \end{tabular}
  }
    \vspace{-1mm}

\end{table}

\begin{table}[!t]
  \centering
   \setlength{\tabcolsep}{10pt}
  \caption{Ablation study of RFR and $\mathcal{L}_\text{AP}$}
  \label{tab:ablation_main}
  \resizebox{0.8\linewidth}{!}{
    \begin{tabular}{l|cccccc}
      \toprule[1.5pt]
      Method 
      & PSNR$\uparrow$ 
      & LPIPS$\downarrow$ 
      & MANIQA$\uparrow$ 
      & MUSIQ$\uparrow$ 
      & CLIPIQA$\uparrow$ 
      & TOPIQ\_IAA$\uparrow$ \\
      \midrule
      
      Base 
      & \textbf{25.32} & \textbf{0.2976} & 0.3691 & 52.94 & 0.5601 & 4.1580 \\
      
      Base$_{\text{w/RFR}}$ 
      & 24.64 & \underline{0.3244} & \underline{0.3908} & 54.73 & 0.5869 & \underline{4.3313} \\
      
      Base$_{\text{w/$\mathcal{L}_\mathbf{AP}$}}$ 
      & \underline{24.85} & 0.3150 & 0.3780 & \underline{55.12} & \underline{0.5920} & 4.2450 \\
      
      Base$_{\text{w/RFR\& $\mathcal{L}_\mathbf{AP}$ }}$(Ours) 
      & 24.31 & 0.3297 & \textbf{0.3935} & \textbf{56.51} & \textbf{0.6135} & \textbf{4.3747} \\
      
      \bottomrule[1.5pt]
    \end{tabular}
  }
    \vspace{-3mm}
\end{table}

 \noindent \textbf{Qualitative Comparisons.}  
 Fig.~\ref{fig:syn} presents the visual comparisons on the synthetic datasets. For the example from the 4KSR-Syn dataset, we observe that certain methods relying on online text extraction and patch-based inference suffer from limited receptive fields. This restriction leads to severe semantic misinterpretations during the SR process, such as erroneously synthesizing streetlamps as leaf textures. Additionally, the results generated by OSEDiff exhibit distinct color inconsistencies and stitching artifacts at the patch fusion boundaries. Meanwhile, OMGSR, a Flux-based one-step SR method, suffers from noticeable periodic artifacts. The example from the DIV4K dataset further substantiates the significant advantages of our approach in suppressing global artifacts and preventing semantic confusion. Fig.~\ref{fig:real} illustrates the visual comparisons under real-world degradation scenarios. As demonstrated, when dealing with complex real-world images, our method stands out as the only model capable of accurately reconstructing regular tile structures and sharp text edges.

\noindent \textbf{Evaluation of efficiency.}
Inference efficiency is evaluated on an NVIDIA H20 GPU with $1024 \times 1024$ inputs and $4096 \times 4096$ outputs. As shown in Table~\ref{tab:efficiency}, OP4KSR requires only 5.75s per image, approximately $16\times$ faster than Flux-based one-step diffusion method OMGSR. Compared to multi-step SUPIR, which requires 1643.7s per image, our method is approximately $280\times$ faster while maintaining competitive memory usage of 32.88G.

\subsection{Ablation}

\textbf{Ablation for Periodic Artifact Suppression Strategy.}
Table \ref{tab:ablation_main} verifies the effectiveness of our periodic artifact suppression strategy. 
Although the Base model achieves high PSNR, it suffers from severe periodic artifacts and limited perceptual quality. 
Introducing $\mathcal{L}_{\text{AP}}$ effectively suppresses grid anomalies, while RFR enhances structural fidelity by alleviating phase collapse. As they address complementary bottlenecks, their combination yields the best perceptual performance, confirming the necessity of both components.

\begin{figure*}[!t]
  \centering
\vspace{-1mm}
  \includegraphics[width=1\linewidth]{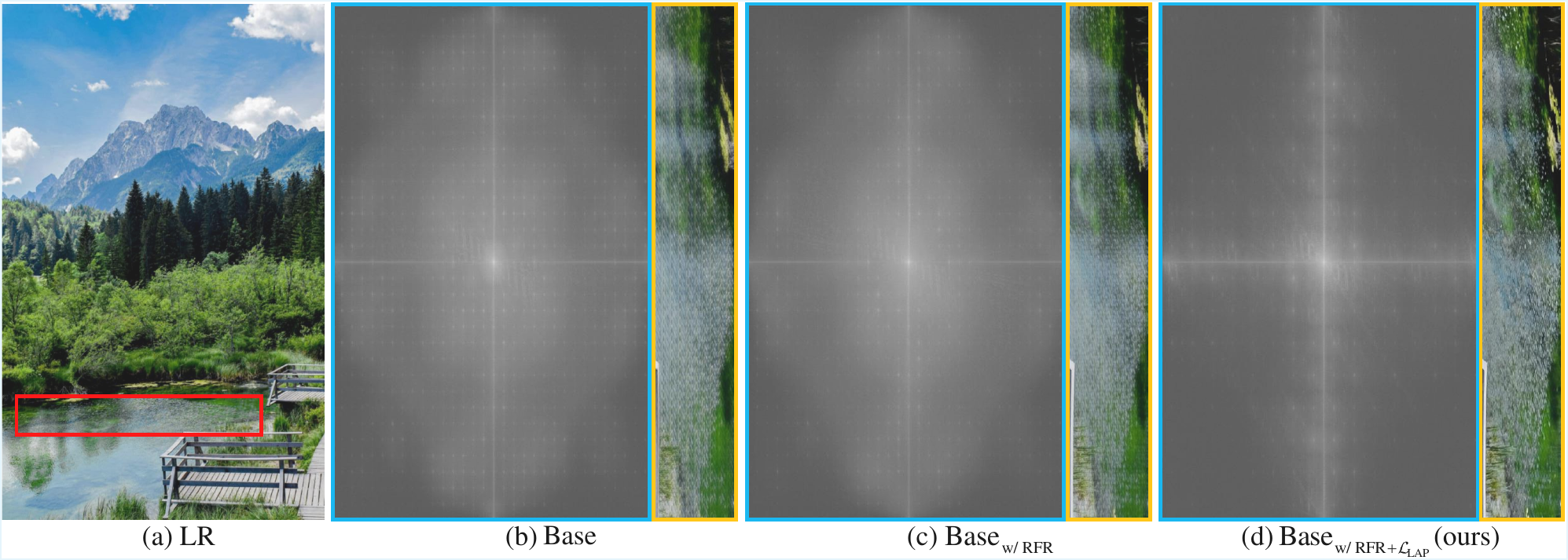}
\caption{ Ablation study of the proposed artifact suppression mechanism in frequency and spatial domains. Blue boxes show the 2D FFT spectra of the full super-resolved images, while yellow boxes highlight local spatial details corresponding to the red box in (a). We compare: (a) the LR input, (b) baseline, (c) baseline + RFR, and (d) our full method (RFR + $\mathcal{L}_{\text{AP}}$).}
  \vspace{-3mm}
  \label{fig:fft}
\end{figure*}

To further validates the effectiveness of our proposed artifact suppression mechanism, we conduct a 2D FFT and spatial analysis in Fig.~\ref{fig:fft}. As shown in Fig.~\ref{fig:fft}(b), the baseline spectrum exhibits anomalous, grid-like high-frequency spikes, corresponding to the conspicuous woven-like periodic artifacts in the spatial domain. With RoPE base frequency rescaling (Fig.~\ref{fig:fft}(c)), this abnormal energy concentration is partially diluted, yet the underlying periodic structures remain. Finally, synergizing RFR with our autocorrelatio-based periodicity  loss (Fig.~\ref{fig:fft}(d)) successfully suppresses these anomalous spikes. The spectrum is restored to a smooth distribution, and spatial grid artifacts are suppressed.

\section{Conclusion}
This paper introduces OP4KSR, a patch-free one-step method for 4K super-resolution. Equipped with an F16 high-compression VAE, our method achieves exceptionally fast inference speeds and facilitates direct 4K outputs. To tackle the inherent periodic artifacts, we explicitly rectify the base frequency of 2D-RoPE and introduce a novel autocorrelation-based periodicity loss, resulting in substantial artifact suppression. Additionally, to propel the development of the 4K super-resolution task, we curate the 4KSR-Train dataset and establish three benchmarks, effectively addressing the current scarcity of dedicated training datasets and real-world benchmarks in the 4K super-resolution field.
 
\section{Limitation} Despite the competitive performance achieved by OP4KSR, our method still exhibits limitations in recovering fine details. For future work, we plan to introduce a high-frequency compensation mechanism to mitigate the physical truncation caused by the extreme compression ratio, thereby further enhancing the model's ability to reconstruct complex textures.
 
 
\clearpage  

%
%
\bibliographystyle{splncs04}
\bibliography{main}

@String(AAAI  = {AAAI})

@String(TOG   = {ACM Trans. Graph.})

@String(TOG   = {ACM TOG})

@article{IHMambaSR,
title = {IHMambaSR: An importance-guided hierarchical mamba with dynamic prompt for single image super-resolution},
journal = {Pattern Recognition},
volume = {175},
pages = {113057},
year = {2026},
issn = {0031-3203},
author = {Chengyan Deng and Kai Zhang and Lieqiang Yang and Wang Zhang and Li Yu},
}

@inproceedings{swinir,
  title={Swinir: Image restoration using swin transformer},
  author={Liang, Jingyun and Cao, Jiezhang and Sun, Guolei and Zhang, Kai and Van Gool, Luc and Timofte, Radu},
  booktitle={Proceedings of the IEEE/CVF international conference on computer vision},
  pages={1833--1844},
  year={2021}
}

@inproceedings{mambair,
  title={Mambair: A simple baseline for image restoration with state-space model},
  author={Guo, Hang and Li, Jinmin and Dai, Tao and Ouyang, Zhihao and Ren, Xudong and Xia, Shu-Tao},
  booktitle={European conference on computer vision},
  pages={222--241},
  year={2024},
  organization={Springer}
}

@inproceedings{diffbir,
  title={Diffbir: Toward blind image restoration with generative diffusion prior},
  author={Lin, Xinqi and He, Jingwen and Chen, Ziyan and Lyu, Zhaoyang and Dai, Bo and Yu, Fanghua and Qiao, Yu and Ouyang, Wanli and Dong, Chao},
  booktitle={European conference on computer vision},
  pages={430--448},
  year={2024},
  organization={Springer}
}

@inproceedings{PASD,
  title={Pixel-aware stable diffusion for realistic image super-resolution and personalized stylization},
  author={Yang, Tao and Wu, Rongyuan and Ren, Peiran and Xie, Xuansong and Zhang, Lei},
  booktitle={European conference on computer vision},
  pages={74--91},
  year={2024},
  organization={Springer}
}

@inproceedings{faithdiff,
  title={Faithdiff: Unleashing diffusion priors for faithful image super-resolution},
  author={Chen, Junyang and Pan, Jinshan and Dong, Jiangxin},
  booktitle={Proceedings of the Computer Vision and Pattern Recognition Conference},
  pages={28188--28197},
  year={2025}
}

@inproceedings{tsdsr,
  title={Tsd-sr: One-step diffusion with target score distillation for real-world image super-resolution},
  author={Dong, Linwei and Fan, Qingnan and Guo, Yihong and Wang, Zhonghao and Zhang, Qi and Chen, Jinwei and Luo, Yawei and Zou, Changqing},
  booktitle={Proceedings of the Computer Vision and Pattern Recognition Conference},
  pages={23174--23184},
  year={2025}
}

@article{resshift,
  title={Resshift: Efficient diffusion model for image super-resolution by residual shifting},
  author={Yue, Zongsheng and Wang, Jianyi and Loy, Chen Change},
  journal={Advances in Neural Information Processing Systems},
  volume={36},
  pages={13294--13307},
  year={2023}
}

@inproceedings{sinsr,
  title={Sinsr: diffusion-based image super-resolution in a single step},
  author={Wang, Yufei and Yang, Wenhan and Chen, Xinyuan and Wang, Yaohui and Guo, Lanqing and Chau, Lap-Pui and Liu, Ziwei and Qiao, Yu and Kot, Alex C and Wen, Bihan},
  booktitle={Proceedings of the IEEE/CVF conference on computer vision and pattern recognition},
  pages={25796--25805},
  year={2024}
}

@misc{flux,
author={Black Forest Labs},
title={FLUX},
year={2023},
howpublished={\url{https://github.com/black-forest-labs/flux}},
}

@article{sdxl,
title={Sdxl: Improving latent diffusion models for high-resolution image synthesis},
author={Podell, Dustin and English, Zion and Lacey, Kyle and Blattmann, Andreas and Dockhorn, Tim and M{\"u}ller, Jonas and Penna, Joe and Rombach, Robin},
journal={arXiv preprint arXiv:2307.01952},
year={2023}
}

@article{vsdloss,
  title={Prolificdreamer: High-fidelity and diverse text-to-3d generation with variational score distillation},
  author={Wang, Zhengyi and Lu, Cheng and Wang, Yikai and Bao, Fan and Li, Chongxuan and Su, Hang and Zhu, Jun},
  journal={Advances in neural information processing systems},
  volume={36},
  pages={8406--8441},
  year={2023}
}

@inproceedings{pisasr,
  title={Pixel-level and semantic-level adjustable super-resolution: A dual-lora approach},
  author={Sun, Lingchen and Wu, Rongyuan and Ma, Zhiyuan and Liu, Shuaizheng and Yi, Qiaosi and Zhang, Lei},
  booktitle={Proceedings of the Computer Vision and Pattern Recognition Conference},
  pages={2333--2343},
  year={2025}
}

@article{osediff,
  title={One-step effective diffusion network for real-world image super-resolution},
  author={Wu, Rongyuan and Sun, Lingchen and Ma, Zhiyuan and Zhang, Lei},
  journal={Advances in Neural Information Processing Systems},
  volume={37},
  pages={92529--92553},
  year={2024}
}

@article{ctmsr,
  title={Consistency Trajectory Matching for One-Step Generative Super-Resolution},
  author={You, Weiyi and Zhang, Mingyang and Zhang, Leheng and Zhou, Xingyu and Shi, Kexuan and Gu, Shuhang},
  journal={arXiv preprint arXiv:2503.20349},
  year={2025}
}

@inproceedings{srformer,
  title={Srformer: Permuted self-attention for single image super-resolution},
  author={Zhou, Yupeng and Li, Zhen and Guo, Chun-Le and Bai, Song and Cheng, Ming-Ming and Hou, Qibin},
  booktitle={Proceedings of the IEEE/CVF International Conference on Computer Vision},
  pages={12780--12791},
  year={2023}
}

@inproceedings{bsrgan,
  title={Designing a practical degradation model for deep blind image super-resolution},
  author={Zhang, Kai and Liang, Jingyun and Van Gool, Luc and Timofte, Radu},
  booktitle={Proceedings of the IEEE/CVF international conference on computer vision},
  pages={4791--4800},
  year={2021}
}

@inproceedings{realesrgan,
  title={Real-esrgan: Training real-world blind super-resolution with pure synthetic data},
  author={Wang, Xintao and Xie, Liangbin and Dong, Chao and Shan, Ying},
  booktitle={Proceedings of the IEEE/CVF international conference on computer vision},
  pages={1905--1914},
  year={2021}
}

@article{omgsr,
  title={OMGSR: You Only Need One Mid-timestep Guidance for Real-World Image Super-Resolution},
  author={Wu, Zhiqiang and Sun, Zhaomang and Zhou, Tong and Fu, Bingtao and Cong, Ji and Dong, Yitong and Zhang, Huaqi and Tang, Xuan and Chen, Mingsong and Wei, Xian},
  journal={arXiv preprint arXiv:2508.08227},
  year={2025}
}

@article{lucidflux,
  title={LucidFlux: Caption-Free Universal Image Restoration via a Large-Scale Diffusion Transformer},
  author={Fei, Song and Ye, Tian and Wang, Lujia and Zhu, Lei},
  journal={arXiv preprint arXiv:2509.22414},
  year={2025}
}

@article{fluxsr,
  title={One diffusion step to real-world super-resolution via flow trajectory distillation},
  author={Li, Jianze and Cao, Jiezhang and Guo, Yong and Li, Wenbo and Zhang, Yulun},
  journal={arXiv preprint arXiv:2502.01993},
  year={2025}
}

@inproceedings{seesr,
  title={Seesr: Towards semantics-aware real-world image super-resolution},
  author={Wu, Rongyuan and Yang, Tao and Sun, Lingchen and Zhang, Zhengqiang and Li, Shuai and Zhang, Lei},
  booktitle={Proceedings of the IEEE/CVF conference on computer vision and pattern recognition},
  pages={25456--25467},
  year={2024}
}

@article{addsr,
  title={Addsr: Accelerating diffusion-based blind super-resolution with adversarial diffusion distillation},
  author={Tai, Ying and Xie, Rui and Zhao, Chen and Zhang, Kai and Zhang, Zhenyu and Zhou, Jun and Yang, Jian},
  journal={Pattern Recognition},
  pages={113012},
  year={2026},
  publisher={Elsevier}
}

@inproceedings{supir,
  title={Scaling up to excellence: Practicing model scaling for photo-realistic image restoration in the wild},
  author={Yu, Fanghua and Gu, Jinjin and Li, Zheyuan and Hu, Jinfan and Kong, Xiangtao and Wang, Xintao and He, Jingwen and Qiao, Yu and Dong, Chao},
  booktitle={Proceedings of the IEEE/CVF conference on computer vision and pattern recognition},
  pages={25669--25680},
  year={2024}
}

@inproceedings{upsr,
  title={Uncertainty-guided Perturbation for Image Super-Resolution Diffusion Model},
  author={Zhang, Leheng and You, Weiyi and Shi, Kexuan and Gu, Shuhang},
  booktitle={Proceedings of the Computer Vision and Pattern Recognition Conference},
  pages={17980--17989},
  year={2025}
}

@article{wang2004image,
title={Image quality assessment: from error visibility to structural similarity},
author={Wang, Zhou and Bovik, Alan C and Sheikh, Hamid R and Simoncelli, Eero P},
journal={IEEE transactions on image processing},
volume={13},
number={4},
pages={600--612},
year={2004},
publisher={IEEE}
}

@article{zhang2015feature,
title={A feature-enriched completely blind image quality evaluator},
author={Zhang, Lin and Zhang, Lei and Bovik, Alan C},
journal={IEEE Transactions on Image Processing},
volume={24},
number={8},
pages={2579--2591},
year={2015},
publisher={IEEE}
}

@inproceedings{zhang2018unreasonable,
title={The unreasonable effectiveness of deep features as a perceptual metric},
author={Zhang, Richard and Isola, Phillip and Efros, Alexei A and Shechtman, Eli and Wang, Oliver},
booktitle={Proceedings of the IEEE conference on computer vision and pattern recognition},
pages={586--595},
year={2018}
}

@inproceedings{yang2022maniqa,
title={Maniqa: Multi-dimension attention network for no-reference image quality assessment},
author={Yang, Sidi and Wu, Tianhe and Shi, Shuwei and Lao, Shanshan and Gong, Yuan and Cao, Mingdeng and Wang, Jiahao and Yang, Yujiu},
booktitle={Proceedings of the IEEE/CVF Conference on Computer Vision and Pattern Recognition},
pages={1191--1200},
year={2022}
}

@inproceedings{ke2021musiq,
title={Musiq: Multi-scale image quality transformer},
author={Ke, Junjie and Wang, Qifei and Wang, Yilin and Milanfar, Peyman and Yang, Feng},
booktitle={Proceedings of the IEEE/CVF international conference on computer vision},
pages={5148--5157},
year={2021}
}

@inproceedings{wang2023exploring,
title={Exploring clip for assessing the look and feel of images},
author={Wang, Jianyi and Chan, Kelvin CK and Loy, Chen Change},
booktitle={Proceedings of the AAAI Conference on Artificial Intelligence},
volume={37},
number={2},
pages={2555--2563},
year={2023}
}

@article{chen2024topiq,
  title={Topiq: A top-down approach from semantics to distortions for image quality assessment},
  author={Chen, Chaofeng and Mo, Jiadi and Hou, Jingwen and Wu, Haoning and Liao, Liang and Sun, Wenxiu and Yan, Qiong and Lin, Weisi},
  journal={IEEE Transactions on Image Processing},
  volume={33},
  pages={2404--2418},
  year={2024},
  publisher={IEEE}
}

@article{stablesr,
  title={Exploiting diffusion prior for real-world image super-resolution},
  author={Wang, Jianyi and Yue, Zongsheng and Zhou, Shangchen and Chan, Kelvin CK and Loy, Chen Change},
  journal={International Journal of Computer Vision},
  volume={132},
  number={12},
  pages={5929--5949},
  year={2024},
  publisher={Springer}
}

@article{schuhmann2022laion,
  title={Laion-5b: An open large-scale dataset for training next generation image-text models},
  author={Schuhmann, Christoph and Beaumont, Romain and Vencu, Richard and Gordon, Cade and Wightman, Ross and Cherti, Mehdi and Coombes, Theo and Katta, Aarush and Mullis, Clayton and Wortsman, Mitchell and others},
  journal={Advances in neural information processing systems},
  volume={35},
  pages={25278--25294},
  year={2022}
}

@article{blattmann2023stable,
  title={Stable video diffusion: Scaling latent video diffusion models to large datasets},
  author={Blattmann, Andreas and Dockhorn, Tim and Kulal, Sumith and Mendelevitch, Daniel and Kilian, Maciej and Lorenz, Dominik and Levi, Yam and English, Zion and Voleti, Vikram and Letts, Adam and others},
  journal={arXiv preprint arXiv:2311.15127},
  year={2023}
}

@inproceedings{invsr,
  title={Arbitrary-steps image super-resolution via diffusion inversion},
  author={Yue, Zongsheng and Liao, Kang and Loy, Chen Change},
  booktitle={Proceedings of the Computer Vision and Pattern Recognition Conference},
  pages={23153--23163},
  year={2025}
}

@article{hypir,
  title={Harnessing diffusion-yielded score priors for image restoration},
  author={Lin, Xinqi and Yu, Fanghua and Hu, Jinfan and You, Zhiyuan and Shi, Wu and Ren, Jimmy S and Gu, Jinjin and Dong, Chao},
  journal={ACM Transactions on Graphics (TOG)},
  volume={44},
  number={6},
  pages={1--21},
  year={2025},
  publisher={ACM New York, NY, USA}
}

@inproceedings{dit4sr,
  title={Dit4sr: Taming diffusion transformer for real-world image super-resolution},
  author={Duan, Zheng-Peng and Zhang, Jiawei and Jin, Xin and Zhang, Ziheng and Xiong, Zheng and Zou, Dongqing and Ren, Jimmy S and Guo, Chunle and Li, Chongyi},
  booktitle={Proceedings of the IEEE/CVF International Conference on Computer Vision},
  pages={18948--18958},
  year={2025}
}

@article{srcnn,
  title={Image super-resolution using deep convolutional networks},
  author={Dong, Chao and Loy, Chen Change and He, Kaiming and Tang, Xiaoou},
  journal={IEEE transactions on pattern analysis and machine intelligence},
  volume={38},
  number={2},
  pages={295--307},
  year={2015},
  publisher={IEEE}
}

@inproceedings{EDSR,
  title={Enhanced deep residual networks for single image super-resolution},
  author={Lim, Bee and Son, Sanghyun and Kim, Heewon and Nah, Seungjun and Mu Lee, Kyoung},
  booktitle={Proceedings of the IEEE conference on computer vision and pattern recognition workshops},
  pages={136--144},
  year={2017}
}

@inproceedings{SAN,
  title={Second-order attention network for single image super-resolution},
  author={Dai, Tao and Cai, Jianrui and Zhang, Yongbing and Xia, Shu-Tao and Zhang, Lei},
  booktitle={Proceedings of the IEEE/CVF conference on computer vision and pattern recognition},
  pages={11065--11074},
  year={2019}
}

@inproceedings{nlsa,
  title={Image super-resolution with non-local sparse attention},
  author={Mei, Yiqun and Fan, Yuchen and Zhou, Yuqian},
  booktitle={Proceedings of the IEEE/CVF conference on computer vision and pattern recognition},
  pages={3517--3526},
  year={2021}
}

@inproceedings{ATD,
  title={Transcending the limit of local window: Advanced super-resolution transformer with adaptive token dictionary},
  author={Zhang, Leheng and Li, Yawei and Zhou, Xingyu and Zhao, Xiaorui and Gu, Shuhang},
  booktitle={Proceedings of the IEEE/CVF Conference on Computer Vision and Pattern Recognition},
  pages={2856--2865},
  year={2024}
}

@inproceedings{PFT,
  title={Progressive Focused Transformer for Single Image Super-Resolution},
  author={Long, Wei and Zhou, Xingyu and Zhang, Leheng and Gu, Shuhang},
  booktitle={Proceedings of the Computer Vision and Pattern Recognition Conference},
  pages={2279--2288},
  year={2025}
}

@inproceedings{mair,
  title={Mair: A locality-and continuity-preserving mamba for image restoration},
  author={Li, Boyun and Zhao, Haiyu and Wang, Wenxin and Hu, Peng and Gou, Yuanbiao and Peng, Xi},
  booktitle={Proceedings of the Computer Vision and Pattern Recognition Conference},
  pages={7491--7501},
  year={2025}
}

@article{tamambair,
  title={Directing Mamba to Complex Textures: An Efficient Texture-Aware State Space Model for Image Restoration},
  author={Peng, Long and Di, Xin and Feng, Zhanfeng and Li, Wenbo and Pei, Renjing and Wang, Yang and Fu, Xueyang and Cao, Yang and Zha, Zheng-Jun},
  journal={arXiv preprint arXiv:2501.16583},
  year={2025}
}

@inproceedings{dasr,
  title={Unsupervised real-world image super resolution via domain-distance aware training},
  author={Wei, Yunxuan and Gu, Shuhang and Li, Yawei and Timofte, Radu and Jin, Longcun and Song, Hengjie},
  booktitle={Proceedings of the IEEE/CVF conference on computer vision and pattern recognition},
  pages={13385--13394},
  year={2021}
}

@inproceedings{srgan,
  title={Photo-realistic single image super-resolution using a generative adversarial network},
  author={Ledig, Christian and Theis, Lucas and Husz{\'a}r, Ferenc and Caballero, Jose and Cunningham, Andrew and Acosta, Alejandro and Aitken, Andrew and Tejani, Alykhan and Totz, Johannes and Wang, Zehan and others},
  booktitle={Proceedings of the IEEE conference on computer vision and pattern recognition},
  pages={4681--4690},
  year={2017}
}

@inproceedings{diffusion4k,
  title={Diffusion-4k: Ultra-high-resolution image synthesis with latent diffusion models},
  author={Zhang, Jinjin and Huang, Qiuyu and Liu, Junjie and Guo, Xiefan and Huang, Di},
  booktitle={Proceedings of the Computer Vision and Pattern Recognition Conference},
  pages={23464--23473},
  year={2025}
}

@article{ultraflux,
  title={UltraFlux: Data-Model Co-Design for High-quality Native 4K Text-to-Image Generation across Diverse Aspect Ratios},
  author={Ye, Tian and Fei, Song and Zhu, Lei},
  journal={arXiv preprint arXiv:2511.18050},
  year={2025}
}

@inproceedings{chen2024pixart,
  title={Pixart-$\sigma$: Weak-to-strong training of diffusion transformer for 4k text-to-image generation},
  author={Chen, Junsong and Ge, Chongjian and Xie, Enze and Wu, Yue and Yao, Lewei and Ren, Xiaozhe and Wang, Zhongdao and Luo, Ping and Lu, Huchuan and Li, Zhenguo},
  booktitle={European Conference on Computer Vision},
  pages={74--91},
  year={2024},
  organization={Springer}
}

@inproceedings{adcsr,
  title={Adversarial diffusion compression for real-world image super-resolution},
  author={Chen, Bin and Li, Gehui and Wu, Rongyuan and Zhang, Xindong and Chen, Jie and Zhang, Jian and Zhang, Lei},
  booktitle={Proceedings of the IEEE/CVF conference on computer vision and pattern recognition},
  pages={28208--28220},
  year={2025}
}

@article{chen2025bridging,
  title={Bridging Fidelity-Reality with Controllable One-Step Diffusion for Image Super-Resolution},
  author={Chen, Hao and Chen, Junyang and Pan, Jinshan and Dong, Jiangxin},
  journal={arXiv preprint arXiv:2512.14061},
  year={2025}
}

@article{rcoe,
  title={Realism Control One-step Diffusion for Real-World Image Super-Resolution},
  author={Wu, Zongliang and Zheng, Siming and Jiang, Peng-Tao and Yuan, Xin},
  journal={arXiv preprint arXiv:2509.10122},
  year={2025}
}

@article{s3diff,
  title={Degradation-guided one-step image super-resolution with diffusion priors},
  author={Zhang, Aiping and Yue, Zongsheng and Pei, Renjing and Ren, Wenqi and Cao, Xiaochun},
  journal={arXiv preprint arXiv:2409.17058},
  year={2024}
}

@inproceedings{tvt,
  title={Fine-structure preserved real-world image super-resolution via transfer vae training},
  author={Yi, Qiaosi and Li, Shuai and Wu, Rongyuan and Sun, Lingchen and Wu, Yuhui and Zhang, Lei},
  booktitle={Proceedings of the IEEE/CVF international conference on computer vision},
  pages={12415--12426},
  year={2025}
}

@inproceedings{mou2022metric,
  title={Metric learning based interactive modulation for real-world super-resolution},
  author={Mou, Chong and Wu, Yanze and Wang, Xintao and Dong, Chao and Zhang, Jian and Shan, Ying},
  booktitle={European conference on computer vision},
  pages={723--740},
  year={2022},
  organization={Springer}
}

@article{bai2025qwen3,
  title={Qwen3-vl technical report},
  author={Bai, Shuai and Cai, Yuxuan and Chen, Ruizhe and Chen, Keqin and Chen, Xionghui and Cheng, Zesen and Deng, Lianghao and Ding, Wei and Gao, Chang and Ge, Chunjiang and others},
  journal={arXiv preprint arXiv:2511.21631},
  year={2025}
}

@inproceedings{rt4ksr,
  title={Towards real-time 4k image super-resolution},
  author={Zamfir, Eduard and Conde, Marcos V and Timofte, Radu},
  booktitle={Proceedings of the IEEE/CVF Conference on Computer Vision and Pattern Recognition},
  pages={1522--1532},
  year={2023}
}

@inproceedings{LRSRN,
  title={Lightweight real-time image super-resolution network for 4k images},
  author={Gankhuyag, Ganzorig and Yoon, Kihwan and Park, Jinman and Son, Haeng Seon and Min, Kyoungwon},
  booktitle={Proceedings of the IEEE/CVF conference on computer vision and pattern recognition},
  pages={1746--1755},
  year={2023}
}

@inproceedings{casr,
  title={Casr: Efficient cascade network structure with channel aligned method for 4k real-time single image super-resolution},
  author={Yoon, Kihwan and Gankhuyag, Ganzorig and Park, Jinman and Son, Haengseon and Min, Kyoungwon},
  booktitle={Proceedings of the IEEE/CVF Conference on Computer Vision and Pattern Recognition},
  pages={7911--7920},
  year={2024}
}

@inproceedings{lsdir,
  title={Lsdir: A large scale dataset for image restoration},
  author={Li, Yawei and Zhang, Kai and Liang, Jingyun and Cao, Jiezhang and Liu, Ce and Gong, Rui and Zhang, Yulun and Tang, Hao and Liu, Yun and Demandolx, Denis and others},
  booktitle={Proceedings of the IEEE/CVF Conference on Computer Vision and Pattern Recognition},
  pages={1775--1787},
  year={2023}
}

@book{dip,
  title={Digital image processing},
  author={J{\"a}hne, Bernd},
  year={2005},
  publisher={Springer}
}

@article{pocketsr,
  title={PocketSR: The Super-Resolution Expert in Your Pocket Mobiles},
  author={Sun, Haoze and Jiang, Linfeng and Li, Fan and Pei, Renjing and Wang, Zhixin and Guo, Yong and Xu, Jiaqi and Chen, Haoyu and Han, Jin and Song, Fenglong and others},
  journal={arXiv preprint arXiv:2510.03012},
  year={2025}
}

@article{zuo20254kagent,
  title={4kagent: agentic any image to 4k super-resolution},
  author={Zuo, Yushen and Zheng, Qi and Wu, Mingyang and Jiang, Xinrui and Li, Renjie and Wang, Jian and Zhang, Yide and Mai, Gengchen and Wang, Lihong V and Zou, James and others},
  journal={arXiv preprint arXiv:2507.07105},
  year={2025}
}

@article{dong2025tinysr,
  title={TinySR: Pruning Diffusion for Real-World Image Super-Resolution},
  author={Dong, Linwei and Fan, Qingnan and Yu, Yuhang and Zhang, Qi and Chen, Jinwei and Luo, Yawei and Zou, Changqing},
  journal={arXiv preprint arXiv:2508.17434},
  year={2025}
}

@inproceedings{zhu2025passionsr,
  title={Passionsr: Post-training quantization with adaptive scale in one-step diffusion based image super-resolution},
  author={Zhu, Libo and Li, Jianze and Qin, Haotong and Li, Wenbo and Zhang, Yulun and Guo, Yong and Yang, Xiaokang},
  booktitle={Proceedings of the Computer Vision and Pattern Recognition Conference},
  pages={12778--12788},
  year={2025}
}

@article{zhu2024oftsr,
  title={Oftsr: One-step flow for image super-resolution with tunable fidelity-realism trade-offs},
  author={Zhu, Yuanzhi and Wang, Ruiqing and Lu, Shilin and Li, Junnan and Yan, Hanshu and Zhang, Kai},
  journal={arXiv preprint arXiv:2412.09465},
  year={2024}
}

@article{talebi2018nima,
  title={NIMA: Neural image assessment},
  author={Talebi, Hossein and Milanfar, Peyman},
  journal={IEEE transactions on image processing},
  volume={27},
  number={8},
  pages={3998--4011},
  year={2018},
  publisher={IEEE}
}

@article{distst,
  title={Image quality assessment: Unifying structure and texture similarity},
  author={Ding, Keyan and Ma, Kede and Wang, Shiqi and Simoncelli, Eero P},
  journal={IEEE transactions on pattern analysis and machine intelligence},
  volume={44},
  number={5},
  pages={2567--2581},
  year={2020},
  publisher={IEEE}
}

@article{ai2024dreamclear,
  title={Dreamclear: High-capacity real-world image restoration with privacy-safe dataset curation},
  author={Ai, Yuang and Zhou, Xiaoqiang and Huang, Huaibo and Han, Xiaotian and Chen, Zhengyu and You, Quanzeng and Yang, Hongxia},
  journal={Advances in Neural Information Processing Systems},
  volume={37},
  pages={55443--55469},
  year={2024}
}

@inproceedings{div,
  title={Ntire 2017 challenge on single image super-resolution: Dataset and study},
  author={Agustsson, Eirikur and Timofte, Radu},
  booktitle={Proceedings of the IEEE conference on computer vision and pattern recognition workshops},
  pages={126--135},
  year={2017}
}

@inproceedings{flickr,
  title={Enhanced deep residual networks for single image super-resolution},
  author={Lim, Bee and Son, Sanghyun and Kim, Heewon and Nah, Seungjun and Mu Lee, Kyoung},
  booktitle={Proceedings of the IEEE conference on computer vision and pattern recognition workshops},
  pages={136--144},
  year={2017}
}

@article{lipman2022flow,
  title={Flow matching for generative modeling},
  author={Lipman, Yaron and Chen, Ricky TQ and Ben-Hamu, Heli and Nickel, Maximilian and Le, Matt},
  journal={arXiv preprint arXiv:2210.02747},
  year={2022}
}

@inproceedings{controlnet,
  title={Adding conditional control to text-to-image diffusion models},
  author={Zhang, Lvmin and Rao, Anyi and Agrawala, Maneesh},
  booktitle={Proceedings of the IEEE/CVF international conference on computer vision},
  pages={3836--3847},
  year={2023}
}

@article{GTASR,
  title={Joint Geometric and Trajectory Consistency Learning for One-Step Real-World Super-Resolution},
  author={Deng, Chengyan and Chen, Zhangquan and Yu, Li and Zhang, Kai and Zhou, Xue and Zhang, Wang},
  journal={arXiv preprint arXiv:2602.24240},
  year={2026}
}

@article{dongjian1,
author = {Yu, Dongjian and Min, Weiqing and Jin, Xin and Jiang, Qian and Jin, Ying and Jiang, Shuqiang},
title = {Diverse and High-Quality Food Image Generation from Only Food Names},
year = {2025},
issue_date = {May 2025},
publisher = {Association for Computing Machinery},
address = {New York, NY, USA},
volume = {21},
number = {5},
issn = {1551-6857},
journal = {ACM Trans. Multimedia Comput. Commun. Appl.},
month = may,
articleno = {153},
numpages = {22}
}

@article{dongjian2,
  title={Food3D: Text-Driven Customizable 3D Food Generation With Gaussian Splatting},
  author={Yu, Dongjian and Min, Weiqing and Jin, Xin and Jiang, Qian and Yao, Shaowen and Jiang, Shuqiang},
  journal={IEEE Transactions on Image Processing},
  volume={34},
  pages={7290--7304},
  year={2025},
  publisher={IEEE}
}
\end{document}